\documentclass[runningheads]{llncs}
\usepackage[T1]{fontenc}
\usepackage{graphicx}
\usepackage{multirow}
\usepackage[table,xcdraw]{xcolor}
\usepackage{mathbbol}
\usepackage{makecell}
\usepackage{subcaption}
\usepackage{tcolorbox}
\usepackage{float}
\usepackage[pagebackref,breaklinks,colorlinks]{hyperref}
\usepackage{amsmath}
\usepackage{enumitem}
\usepackage{tikz}
\usepackage{booktabs}
\usepackage{orcidlink}
\usepackage{marvosym}

\definecolor{lightblue}{HTML}{DCF2FD}
\definecolor{skyblue}{HTML}{92C5DE}
\definecolor{myblue}{HTML}{6691CD}
\definecolor{cellmicroscopycolor}{HTML}{FBE2D5} 
\definecolor{breastimagingcolor}{HTML}{DAE9F8}  
\definecolor{chestxraycolor}{HTML}{C0F1C8}     
\definecolor{fundoscopycolor}{HTML}{FAF6A5}     
\definecolor{retinaloctcolor}{HTML}{D9D9D9}    

\newtcolorbox{HighlighterBox}[2][]{
    arc=3.8pt,
    left=5.0pt,
    right=5.0pt,
    bottom=2pt,
    top=2pt, 
    colback=skyblue!7.5,
    colframe=skyblue!35,  
    boxrule=0.8pt,
    colbacktitle=skyblue!35,  
    coltitle=myblue!20!black,
    title=\textbf{#2},
    fonttitle=\bfseries,
    #1,
}
\newtcolorbox{highlighterbox}[1][]{
    arc=3.8pt,
    left=5.0pt,
    right=5.0pt,
    bottom=2pt,
    top=2pt, 
    rounded corners,
    boxrule=0.8pt,
    colframe=myblue!25,
    colback=myblue!5,
}

\begin{document}
%
\title{On the Robustness of Medical Vision-Language Models: Are they Truly Generalizable?
}
\titlerunning{$\mathbb{R}$obustMedCLIP}
%
\author{Raza Imam(\Letter)\orcidlink{0000-0003-2007-6267} \and
Rufael Marew\orcidlink{0000-0001-8196-698X}\and
Mohammad Yaqub\orcidlink{0000-0001-6896-1105}}
\authorrunning{Imam et al.}
%
\institute{
Mohamed bin Zayed University of Artificial Intelligence, Abu Dhabi, UAE
\email{\texttt{\{raza.imam, rufael.marew, mohammad.yaqub\}}\normalfont{@mbzuai.ac.ae}
}
}
\maketitle              
\renewcommand{\thefootnote}{}%
\footnotetext{
Dataset and Code is available at: \href{https://github.com/BioMedIA-MBZUAI/RobustMedCLIP}{Github}
}%
\footnotetext{
Accepted at: Medical Image Understanding and Analysis (MIUA) 2025
}%
\footnotetext{
Corresponding Author: Raza Imam \Letter (\texttt{raza.imam}@mbzuai.ac.ae)
}%
\renewcommand{\thefootnote}{\arabic{footnote}}

\begin{abstract}
Medical Vision-Language Models (MVLMs) have achieved \textit{par excellence} generalization in medical image analysis, yet their performance under noisy, corrupted conditions remains largely untested.
Clinical imaging is inherently susceptible to acquisition artifacts and noise; however, existing evaluations predominantly assess generally clean datasets, overlooking robustness---\textit{i.e.}, the model's ability to perform under real-world distortions. 
To address this gap, we first introduce MediMeta-C, a corruption benchmark that systematically applies several perturbations across multiple medical imaging datasets. Combined with MedMNIST-C, this establishes a comprehensive robustness evaluation framework for MVLMs. 
We further propose $\mathbb{R}$obustMedCLIP, a visual encoder adaptation of a pretrained MVLM that incorporates few-shot tuning to enhance resilience against corruptions. Through extensive experiments, we benchmark 5 major MVLMs across 5 medical imaging modalities, revealing that existing models exhibit severe degradation under corruption and struggle with domain-modality tradeoffs. 
Our findings highlight the necessity of diverse training and robust adaptation strategies, demonstrating that efficient low-rank adaptation when paired with few-shot tuning, improves robustness while preserving generalization across modalities.

\end{abstract}
\section{Introduction}
In recent years, Medical Vision-Language Models (MVLMs) have emerged as powerful tools for analyzing medical imaging data by leveraging large-scale multimodal learning \cite{wang2022medclipcontrastivelearningunpaired,zhang2025biomedclipmultimodalbiomedicalfoundation,khattak2024unimedclipunifiedimagetextpretraining}. These models have demonstrated impressive accuracy in zero-shot and few-shot medical image classification, making them promising candidates for real-world deployment. However, despite improvements in generalization accuracy, the \textit{robustness} of MVLMs under real-world distribution shifts, from the point of corruptions, remains largely unexplored. Clinical imaging in practice is often affected by artifacts and noise introduced during acquisition and preprocessing, which can significantly degrade model performance. Existing evaluations \cite{zhao2025clip,chen2024survey,deanda2025benchmarking} predominantly focus on clean datasets, overlooking the impact of such corruptions. Without systematic robustness assessment, the reliability of MVLMs in practical medical scenarios remains uncertain, raising concerns about their safety and effectiveness in clinical decision-making.

Although datasets such as CheXpert~\cite{irvin2019chexpert} and MedMNIST~\cite{chen2021medmnist} are carefully curated to ensure high-quality images through fixed resolutions and rigorous normalization, they fail to capture the range of corruptions and distribution shifts encountered in real-world clinical settings. Inspired by ImageNet-C~\cite{hendrycks2019benchmarking}, MedMNIST-C~\cite{di2024medmnist} was proposed to introduce controlled distortions. However, its reliance on low-resolution MedMNIST data, use of modality-specific corruptions, and disregard for the inherently lower high-frequency content of medical images limit its ability to fully represent authentic imaging challenges~\cite{xu2019systematic}. This highlights the need for a more comprehensive corruption benchmark that effectively evaluates the robustness of MVLMs and other medical AI models.

{\renewcommand{\arraystretch}{1.2}
\begin{table}[t]
\caption{\small Overview of dataset statistics for MediMeta~\cite{woerner2024comprehensive} and MedMNIST~\cite{chen2021medmnist}, covering the common imaging modalities analyzed in this study. \#Val/Test represents the number of validation and test samples. Extended statistics are provided in \textbf{Appendix}.}
\resizebox{\textwidth}{!}{%
\begin{tabular}{llll|lll}
 & \multicolumn{3}{c}{\textcolor{purple}{\textbf{MediMeta}}} & \multicolumn{3}{c}{\textcolor{blue}{\textbf{MedMNIST}}} \\
Modality~$\downarrow$ & Data Name~~ & \#Val/Test~~~~ & Description & Data Name~~~ & \#Val/Test~~~~~ & Description \\ \hline
\multirow{1}{*}{\textcolor{orange}{Cell Microscopy}} & PBC & 1,709/3,149 & Blood cells & BloodMNIST & 1,712/3,421 & Blood cells \\
\multirow{1}{*}{\textcolor{myblue}{Breast Imaging}} & Mammo & 214/326 & Calcifications & BreastMNIST & 78/156 & Breast tumors \\ 
\textcolor{teal}{Chest X-ray} & Pneumonia & 817/624 & Lung infection & PneumoniaMNIST & 524/624 & Lung infection \\ 
\textcolor{brown}{Fundoscopy} & Fundus & 640/640 & Eye diseases & RetinaMNIST & 120/400 & Eye diseases \\ 
\textcolor{olive}{Retinal OCT} & OCT & 16,694/1,000 & Retinal layers & OCTMNIST & 10,832/1,000 & Retinal layers
\end{tabular}}
\label{table:data_descriptions}
\vspace{-0.80cm}
\end{table}}

A critical aspect of robustness evaluation is understanding how different types and severities of corruptions impact MVLM performance across various medical imaging modalities (Table \ref{table:data_descriptions}). Traditional learning models typically rely on extensive dataset curation and domain-specific adaptations to improve resilience \cite{khan2025comprehensive}. In contrast, MVLMs introduced a paradigm where contrastive learning could play a pivotal role in addressing robustness challenges. Given their strong generalization capabilities, it is necessary to investigate whether contrastively learned pretrained MVLMs, when effectively combined with robust adaptation techniques, can mitigate the impact of corruptions and enhance MVLMs reliability. 
These challenges motivate us to answer the following:
\begin{HighlighterBox}{Research Questions}
\small
\begin{enumerate}
    \item[Q1.] Do MVLMs maintain performance when facing visual corruptions?
    \item[Q2.] How do corruption types and their severities affect MVLM generalization across clinical modalities?
    \item[Q3.] Do accurate MVLMs directly correlate with robustness against corruptions and distribution shifts?
    \item[Q4.] How can few-shot sampling be \textit{efficiently} infused with the presence of strong MVLMs to address generalization despite corruptions?
\end{enumerate}
\end{HighlighterBox}

To answer these research questions, we introduce \textbf{MediMeta-C}, a corruption benchmark specifically designed for medical imaging. By combining MediMeta-C with MedMNIST-C \cite{di2024medmnist}, we establish a comprehensive evaluation framework to assess model robustness across multiple imaging modalities. Furthermore, we propose \textbf{$\mathbb{R}$obustMedCLIP}, a novel adaptation of a pretrained MVLM (such as BioMedCLIP \cite{zhang2025biomedclipmultimodalbiomedicalfoundation} and MedCLIP \cite{wang2022medclipcontrastivelearningunpaired}) that incorporates few-shot fine-tuning to enhance performance under corrupted conditions. Through extensive experiments, we benchmark MVLMs against a range of seven corruptions, providing valuable insights into their resilience and adaptability in real-world medical settings.
Overall, our key contributions are summarized as follows:
\begin{enumerate}[leftmargin=*, topsep=0.5em]
    \item \textbf{MediMeta-C Dataset}: A corruption classification benchmark for evaluation that applies 7 systematic perturbations to 5 medical imaging datasets to simulate real-world OOD shifts.
    \item \textbf{Corruption MVLM Benchmarking}: A unified evaluation framework combining MediMeta-C and MedMNIST-C to analyze the robustness of 5 major MVLMs across 5 medical imaging modalities across classification tasks.
    \item \textbf{$\mathbb{R}$obustMedCLIP}: A robust adaptation of pretrained MVLM that integrates efficient few-shot-tuning to enhance visual representations to achieve better generalization and robustness against corruptions.
    \item \textbf{Extensive Evaluations}: A systematic study assessing the impact of various corruption types and severities across multiple MVLMs robustness, while evaluating the true generality of existing MVLMs.
    \item \textbf{Datasets and Code}: We release our benchmark dataset and APIs, promoting standardized robustness evaluation practices in medical AI research.
\end{enumerate}

\begin{figure}[t]
    \centering
    \includegraphics[width=0.95\linewidth]{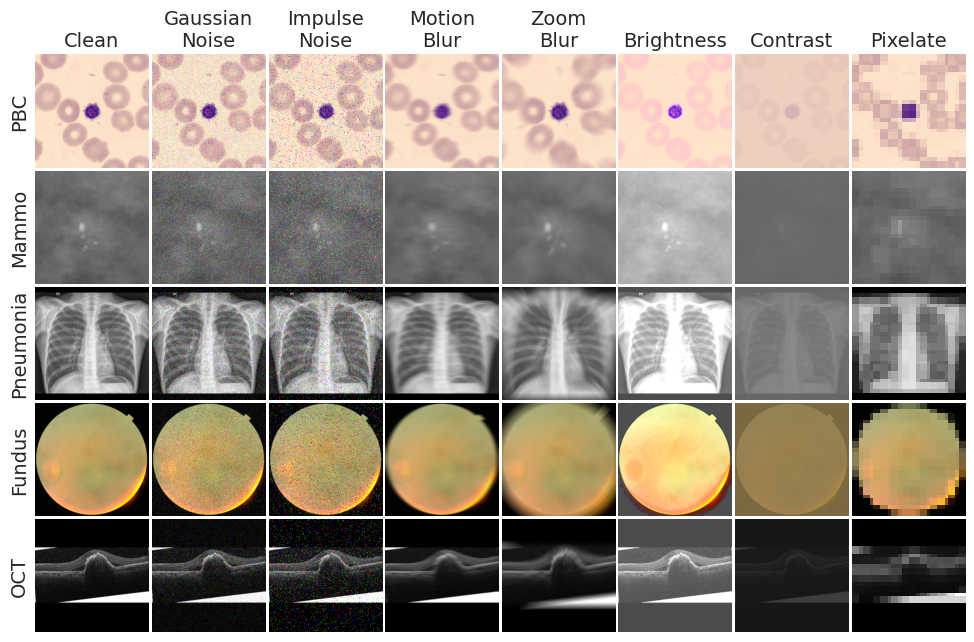}
    \vspace{-0.3cm}
    \caption{\small \textbf{Corrupted samples} from our MediMeta-C dataset. The y-axis shows dataset names by modality and the x-axis displays corruption types at a fixed severity level.}
    \label{fig:medimeta_c}
    \vspace{-0.5cm}
\end{figure}

\begin{figure*}[t]
    \centering
    \begin{minipage}{0.48\textwidth}
        \centering
        \begin{subfigure}[b]{\textwidth}
            \centering
            \includegraphics[width=1.0\linewidth]{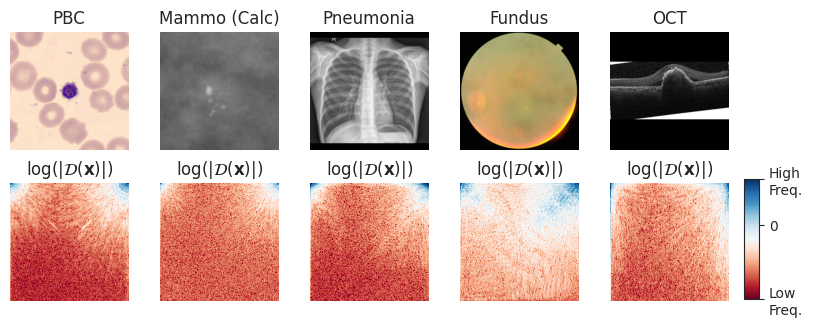}
            \vspace{-0.50cm}
            \caption{\small MediMeta-C DCT frequency}
        \end{subfigure}
        
        \vskip\baselineskip 

        \begin{subfigure}[b]{\textwidth}
            \centering
            \includegraphics[width=1.0\linewidth]{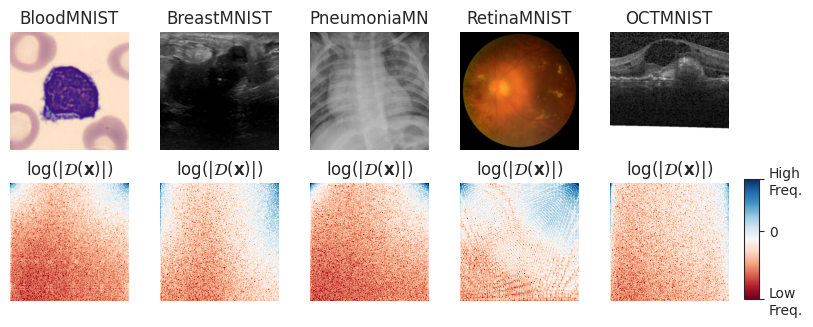}
            \vspace{-0.50cm}
            \caption{\small MedMNIST-C DCT frequency}
        \end{subfigure}
        \vspace{-0.5cm}
        \caption{\small Comparison of average \textbf{DCT frequency distributions} across datasets. Medical images generally exhibit \textit{higher density} of \textcolor{orange}{\textbf{low-frequency content}} compared to natural images and vice-versa~\cite{xu2019systematic}. Among the two, MediMeta-C (a) more clearly demonstrates this assumption than MedMNIST-C (b).}
        \label{fig:dct}
    \end{minipage}%
    \hfill
    \begin{minipage}{0.48\textwidth}
        \centering
        \includegraphics[width=\textwidth]{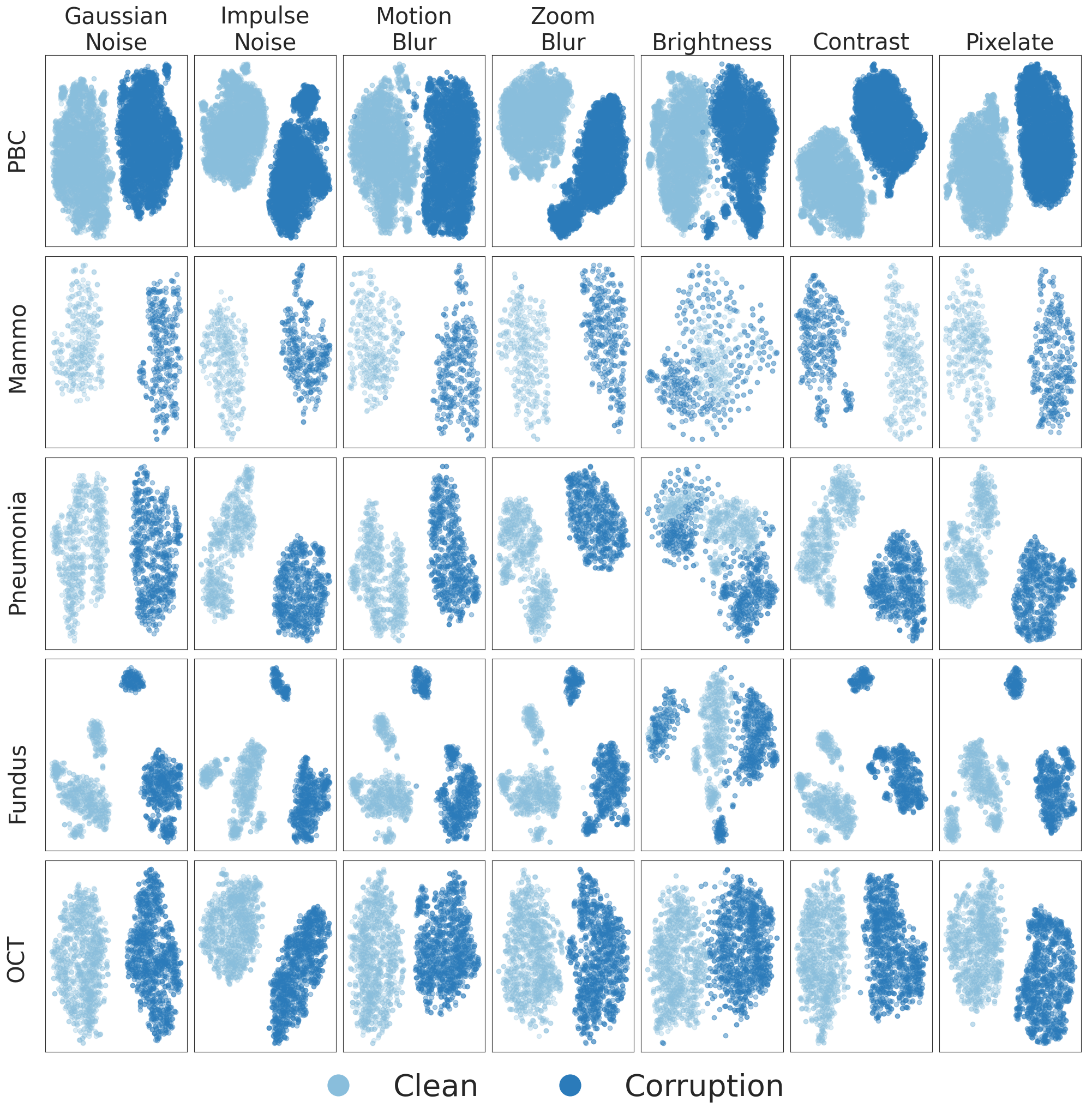}
        \vspace{-0.5cm}
        \caption{\small t-SNE visualization of the \textbf{clean and corrupted feature distributions}, showing how the \textit{distributions shift} occur at the latent-level due to introduced corruption. MediMeta-C's \textcolor[HTML]{2b7bba}{\textbf{Corrupted}} features differ notably than MediMeta's \textcolor[HTML]{89bedc}{\textbf{Clean}} features. Here RN50 backbone is used to extract features.}
        \label{fig:feature_difference}
    \end{minipage}
    \vspace{-0.5cm}
    \end{figure*}

\section{Background}

\noindent\textbf{A. Vision-Language Models in Medical Imaging:}
The adaptation of vision-language models to the medical domain has advanced by modifying dual-encoder architectures. For example, CLIP~\cite{radford2021learning} has been fine-tuned on medical image-text pairs, resulting in variants such as MedCLIP~\cite{wang2022medclipcontrastivelearningunpaired} and BioMedCLIP~\cite{zhang2025biomedclipmultimodalbiomedicalfoundation} that employ domain-specific tokenization and contrastive loss adjustments. However, training on clean, curated datasets leaves these models vulnerable to the distribution shifts and noise present in real-world clinical imaging. This vulnerability underscores the need for corruption-specific adaptations---such as fine-tuning on distorted samples and robust weak or unsupervised strategies \cite{imam2025noise}---to improve resilience against imaging artifacts and ensure reliable performance.

\vspace{0.1cm}
\noindent\textbf{B. Robustness in Healthcare:}
Among the reliable strategies in trustworthy solutions, approaches such as adversarial training \cite{hanif2023frequency,hanif2025frequency,malik2024evaluating} and domain adaptation have been shown to achieve consistent and generalizable model robustness across various recognition tasks \cite{reviewrobustness}. To enhance robustness in healthcare AI, multimodal fusion architectures with knowledge distillation have improved patient outcome predictions by integrating chest X-rays, clinical texts, and tabular data \cite{hayat2022medfuse}. Additionally, combining clinical time-series data with chest X-rays using transformer-based models has boosted diagnostic performance, highlighting the role of multimodal fusion in improving robustness \cite{khader2023multimodal}. These advancements emphasize the need for resilient \cite{hanif2024baple}, multimodal AI systems in healthcare.


\begin{figure}[t]
    \centering
    \includegraphics[width=0.87\linewidth]{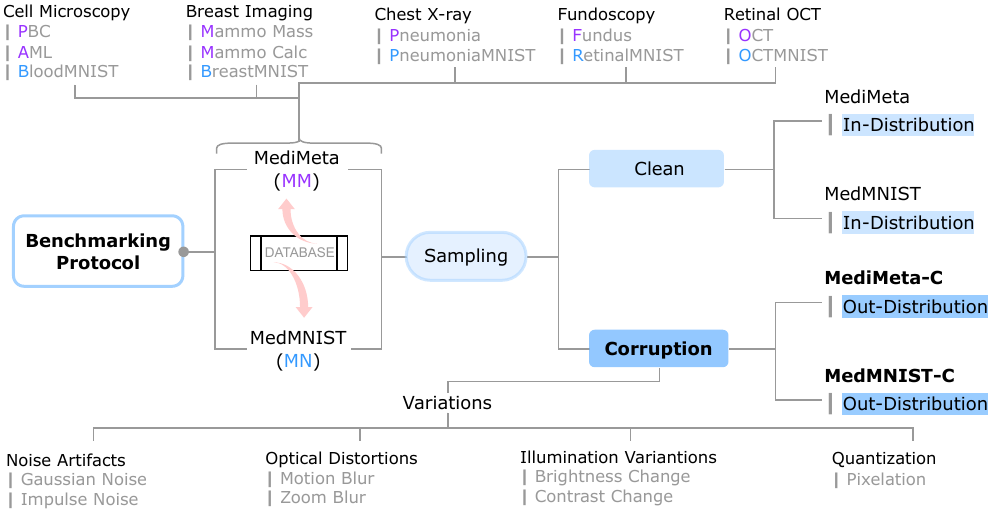}
    \vspace{-0.20cm}
    \caption{\small \textbf{Benchmarking protocol} used in our evaluation, where clean samples represent \textit{In-Distribution} data seen by $\mathbb{R}$MC, while corrupted samples correspond to \textit{Out-Distribution} shifts. \textit{Sampling} refers to selecting the \textit{testset} from each dataset.}
    \label{fig:benchmark_protocol}
    \vspace{-0.5cm}
\end{figure}

\section{Methodology}
\subsection{Medical Corruption Benchmark}

\begin{figure}[t]
    \centering
    \includegraphics[width=1.0\linewidth]{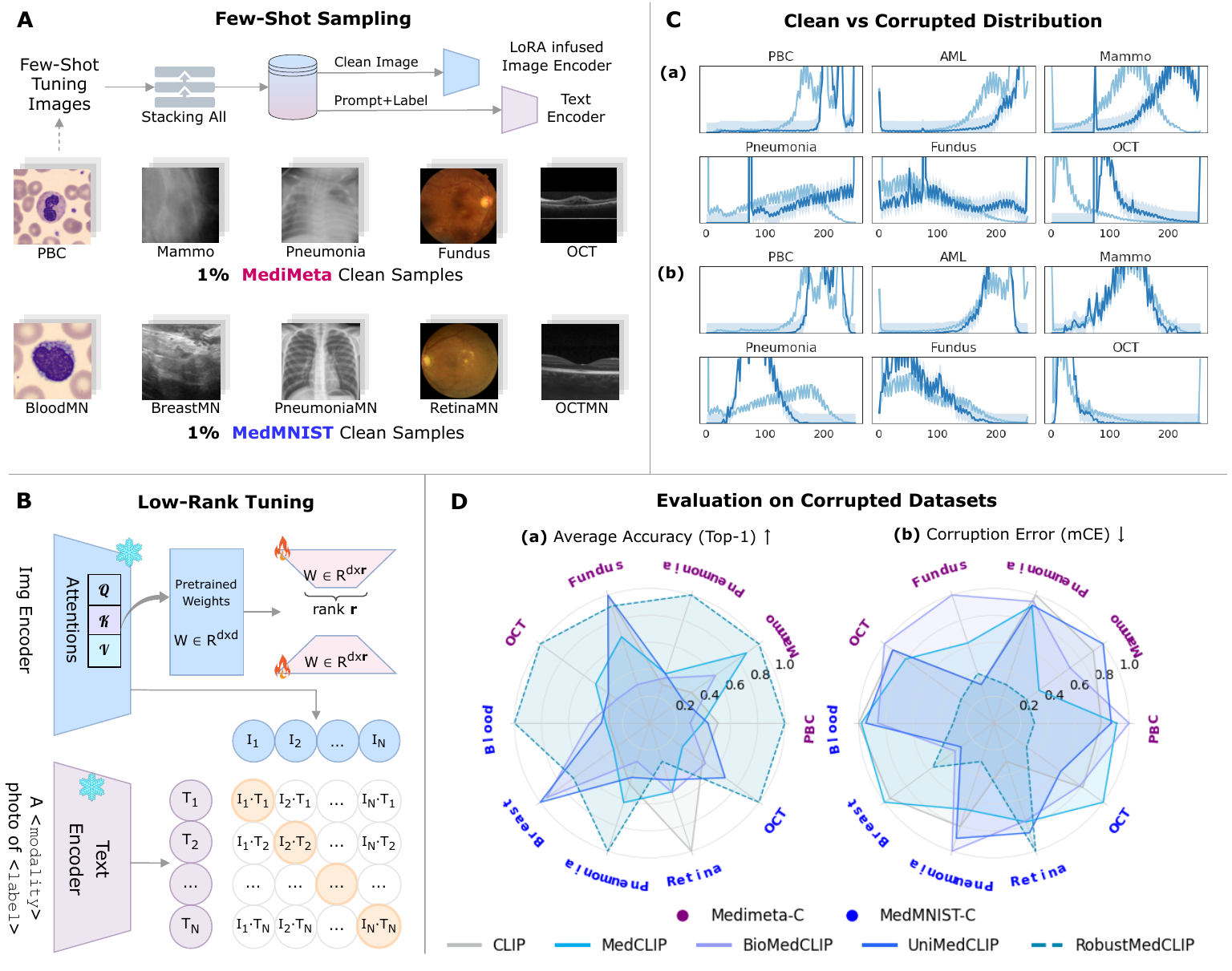}
    \vspace{-0.6cm}
    \caption{\small
    \textbf{A)} Few-shot samples from each modality are drawn from the clean training set to adapt the LoRA-augmented image encoder of the pretrained BioMedCLIP.
    \textbf{B)} Low-rank attention matrices within the image encoder are updated using Eq.~\ref{eq:loss_ft}, enabling the model to learn from diverse in-distribution modalities while retaining pretrained knowledge.
    \textbf{C)} Pixel-level density distributions comparing \textcolor[HTML]{89bedc}{\textbf{Clean}} and \textcolor[HTML]{2b7bba}{\textbf{Corrupted}} samples under (a) brightness and (b) contrast corruptions, highlighting input-level distributional shifts.
    \textbf{D)} (a) Top-1 Accuracy as a measure of \textit{generalization}, and (b) \textit{mean} Corruption Error (mCE) as a proxy for \textit{robustness}, averaged over \textcolor{purple}{MediMeta-C} and \textcolor{blue}{MedMNIST-C}. All values are normalized for visual comparability across models.
    }
    \label{fig:RMedCLIP}
    \vspace{-0.3cm}
\end{figure}

\textbf{A. MediMeta-C Design:}
We introduce MediMeta-C, a corruption benchmark derived from the MediMeta dataset \cite{woerner2024comprehensive} that is designed to emulate distribution shifts encountered in real-world clinical imaging. Our benchmark encompasses 7 distinct corruption types, organized into four primary categories: \textit{Noise Artifacts}, \textit{Optical Distortions}, \textit{Illumination Variations}, and \textit{Quantization} or \textit{Compression} errors depicting real-world medical imaging acquisition and preprocessing errors. To capture the variability depicting real-world corruption, each corruption type is implemented at five severity levels as depicted in Fig.~\ref{fig:medimeta_c}. Unlike MedMNIST-C \cite{di2024medmnist}, which relies on low-resolution data and modality-specific corruptions, MediMeta-C employs diverse perturbations on high-resolution images to better reflect clinical variability. For example, brightness/contrast alterations induce pixel density shifts (Fig.~\ref{fig:RMedCLIP}C), while DCT analysis\footnote{DCT frequency analysis \cite{shen2021dct} applies the Discrete Cosine Transform to convert an image's spatial data into cosine-based frequency components \cite{MAICAS2019101562}.} reveals corrupted images exhibit amplified low-frequency content and suppressed high-frequency signals \cite{xu2019systematic}, a trend more pronounced in MediMeta-C than MedMNIST-C (Fig.~\ref{fig:dct}). Latent feature divergence between clean and corrupted images, visualized via t-SNE (Fig.~\ref{fig:feature_difference}), further confirms the realistic simulation of corruptions in MediMeta-C.

\vspace{0.1cm}
\noindent\textbf{B. Common Medical Distortions:}
Specifically, Noise Artifacts---such as \textit{Gaussian Noise} and \textit{Impulse Noise}---can arise during medical image acquisition due to low-light conditions or sensor bit errors. Optical Distortions, including \textit{Motion Blur} and \textit{Zoom Blur}, often occur when there is patient movement or rapid changes in imaging focus. Illumination Variations, evidenced by shifts in \textit{Brightness} and \textit{Contrast}, are frequently encountered as a result of inconsistent exposure settings or variable ambient lighting during the scanning process. Quantization/Compression errors, like \textit{Pixelation} and JPEG artifacts, may be introduced during image upsampling or through lossy compression techniques used in digital processing. Our benchmark encompasses these seven distinct corruption types to closely simulate the real-world challenges encountered in medical imaging acquisition and preprocessing.

\vspace{0.1cm}
\noindent\textbf{C. Benchmarking Protocol:}
We establish our evaluation framework by leveraging both MediMeta-C and MedMNIST-C to assess the generalization capabilities of MVLMs.  As illustrated in Fig.~\ref{fig:benchmark_protocol}, clean samples from MediMeta and MedMNIST serve as in-distribution (\textit{ID}) data representing instances the model has encountered during training while corrupted samples from corruption datasets correspond to Out-of-Distribution (\textit{OOD}) shifts. 

\begin{HighlighterBox}{Constraint}
\small
    {It is critical to emphasize that our benchmark should be used solely for testing; models should be trained exclusively on clean datasets such as MedMNIST and MediMeta, and not on the corrupted versions provided in MediMeta-C.}
\end{HighlighterBox}
This constraint guarantees that any observed performance degradation is attributable to distribution shifts rather than overfitting to specific corruptions. Comparative evaluation of MediMeta-C and MedMNIST-C underscores the benefits of a corruption benchmark that closely mimics real-world medical imaging distortions, thereby enhancing MVLM reliability. Overall, our MediMeta-C encompasses \textbf{175} (\textit{i.e.}, $7\times5\times5$) distinct corruption sets---derived from 7 corruption types, each applied at 5 severity levels across 5 modality-specific datasets---using \textit{testset} images from MediMeta to rigorously test MVLM generalization.

\subsection{$\mathbb{R}$obustMedCLIP}

$\mathbb{R}$obustMedCLIP (or $\mathbb{R}$MC) enhances robustness against corruption benchmarks by incorporating few-shot fine-tuning into a BioMedCLIP pretrained MVLM. The goal is to efficiently adapt the model using a few clean samples from diverse modalities, achieving improved robustness against corruptions
\footnote{Note that lower \textbf{$\downarrow$} \textit{mean} Corruption Error (mCE) indicates better robustness, while higher \textbf{$\uparrow$} Accuracy reflects stronger generalization. Refer to Section \ref{sec:experimetation} for details.}, 
while retaining the rich, generalizable semantics learned during large-scale pretraining. To achieve this, we update low-rank adapters within the query (\(\mathcal{Q}\)), key (\(\mathcal{K}\)), and value (\(\mathcal{V}\)) matrices of the visual encoder. The training objective focuses on optimizing the image encoder weights using only a limited set of annotated examples. 
\begin{HighlighterBox}{Rationale behind $\mathbb{R}$MC}
\small
    We hypothesize that data-modality diversity is more critical than sheer volume, proposing that adapting a low-ranked image encoder with few-shot tuning across a broad yet representative range of clinical domains---\textit{analogous to training on fewer but highly diverse institutions rather than a large dataset confined to a limited set of hospitals}---enhances robustness while preserving cross-modality generalization.
\end{HighlighterBox}

\noindent\textbf{A. Few-Shot Fine-Tuning:}
Given a dataset \( \mathbb{D} \supset \{(X_i, Y_i)\}_{i=1}^{N} \) consisting of medical image-text pairs (Fig.~\ref{fig:RMedCLIP}A), $\mathbb{R}$MC fine-tunes on a limited subset of annotated examples using contrastively learned pretrained BioMedCLIP. This few-shot tuning allows the MVLM to adapt to unseen distribution shifts without overfitting the training distribution, thereby enhancing its robustness to diverse corruptions and improving generalization. The feature embeddings \(\boldsymbol{f}_v\) and \(\boldsymbol{f}_t\) for the image and text encoders are obtained as:  
\begin{equation}
    \boldsymbol{f}_v = \mathcal{F}_{\theta_v}(X_i), \quad  
    \boldsymbol{f}_t = \mathcal{F}_{\theta_t}(Y_i') \quad \text{where} \quad Y_i'=\langle\texttt{Prompt}\rangle+Y_i,
\end{equation}
here, \( \mathcal{F}_{\theta_v} \) and \( \mathcal{F}_{\theta_t} \) denote the image and text encoders, and \(\langle\texttt{Prompt}\rangle\) is a modality-specific prefix (e.g., \(\langle\texttt{A photo of a}~~\text{modality}(Y_i)\rangle\)). $\mathbb{R}$MC employs cross-entropy between true labels $Y_i$ and zero-shot predictions $\hat{Y}_i$ of pretrained MVLM (such as BioMedCLIP or MedCLIP). The fine-tuning loss \( \mathcal{L}_{\text{FT}} \) that updates $\mathcal{F}_{\theta_v}$ is given as:
\begin{equation}
    \mathcal{L}_{\text{FT}}(\theta_v) = -\sum_{i} Y_i \log \hat{Y}_i \quad \text{where}~ \hat{Y}_i = \text{softmax}(S_{i,c} / \tau) = \frac{\exp(S_{i,c} / \tau)}{\sum_{c'} \exp(S_{i,c'} / \tau)}
    \label{eq:loss_ft},
\end{equation}
with $S_{i,c} = \cos(\boldsymbol{f}_v^i, \boldsymbol{f}_t^c) = \boldsymbol{f}_v^i \cdot \boldsymbol{f}_t^c / (\|\boldsymbol{f}_v^i\|_2 \cdot \|\boldsymbol{f}_t^c\|_2)$ representing the cosine similarity between $\boldsymbol{f}_v^i$ and  $\boldsymbol{f}_t^c$ for class $c$, and $\tau$ being a temperature parameter. 

\vspace{0.2cm}
\noindent\textbf{B. Low-Rank Adapter Optimization:}
To avoid updating all model parameters during fine-tuning, $\mathbb{R}$MC employs \textit{low-rank adaptation} (LoRA) \cite{hu2022lora} to efficiently update only a small subset of parameters in the transformer layers using $\mathcal{L}_{\text{FT}}$. Specifically, the query (\(\mathcal{Q}\)), key (\(\mathcal{K}\)), and value (\(\mathcal{V}\)) matrices are modified via low-rank decompositions as follows:
\begin{equation}
    \mathcal{Q} = \mathcal{Q} + A_{\mathcal{Q}} B_{\mathcal{Q}}, \quad  
    \mathcal{K} = \mathcal{K} + A_{\mathcal{K}} B_{\mathcal{K}}, \quad  
    \mathcal{V} = \mathcal{V} + A_{\mathcal{V}} B_{\mathcal{V}},
\end{equation}
where \( A_{\mathcal{Q}}, A_{\mathcal{K}}, A_{\mathcal{V}} \in \mathbb{R}^{d \times r} \) and \( B_{\mathcal{Q}}, B_{\mathcal{K}}, B_{\mathcal{V}} \in \mathbb{R}^{r \times d} \) are low-rank matrices (with rank \( r \ll d \)). This strategy drastically reduces the number of trainable parameters, enabling efficient adaptation without incurring the heavy computational cost of full fine-tuning \cite{imam2024test}. By restricting updates to the attention layers via LoRA (Fig.~\ref{fig:RMedCLIP}B), $\mathbb{R}$MC efficiently adapts domain-specific corruption patterns while preserving the generalizable representations acquired during pretraining. 

\vspace{0.2cm}
\noindent\textbf{C. Zero-Shot Inference:}
After fine-tuning, $\mathbb{R}$MC performs zero-shot classification by leveraging its robust multimodal representations. Given a test medical image \( X_{\text{test}} \) and a set of textual class descriptions \( T = \{t_1, t_2, \dots, t_c\} \) corresponding to \( c \) categories, the MVLM first encodes the inputs as:
\begin{equation}
    \boldsymbol{f}_v = \mathcal{F}_{\theta_v}(X_{\text{test}}), \quad \boldsymbol{f}_t^i = \mathcal{F}_{\theta_t}(t_i) \quad \text{for } i=1,\dots,c.
\end{equation}
The cosine similarity between the normalized image embedding $\tilde{\boldsymbol{f}}_v$ and each normalized text embedding $\tilde{\boldsymbol{f}}_t$ is computed as $S_{v,t_i}$ using Eq. \ref{eq:loss_ft}:
Finally, the model predicts the class corresponding to the highest similarity:
\begin{equation}
    \hat{y} = \arg\max_{i} S_{v,t_i}.
\end{equation}
This allows the model to assign labels in a zero-shot manner, without requiring any additional task-specific training. 
Moreover, by directly comparing the embeddings in a shared latent space, the MVLM effectively leverages the learned representations to generalize to unseen classes and corrupted conditions (Fig.~\ref{fig:RMedCLIP}D).


\section{Experimentation}
\label{sec:experimetation}

\noindent\textbf{A. Robustness Metric:}
To evaluate MVLM performance under corrupted datasets, we adopt a robustness metric inspired by ImageNet-C. For a given MVLM model \( f \) and a corruption type \( c \) applied at severity levels \( s = 1, \dots, 5 \), the Top-1 error is computed as
\begin{equation}
    E_{s,c}^f = 1 - \text{Acc}_{s,c}^f \quad \text{where~~} \text{Acc}_{s,c}^f \text{~~is the Top-1 accuracy.}
    \label{eq:acc}
\end{equation}
The \textbf{Corruption Error (CE)} for MVLM \( f \) on corruption \( c \) is then defined as
\begin{equation}
    \text{CE}_\text{c}^f = \left( \sum_{s=1}^{5} E_{s,c}^f \right) / \left( \sum_{s=1}^{5} E_{s,c}^{\text{baseline}} \right),
\end{equation}
where the baseline is set to OpenAI CLIP with ViT-B/16. Finally, the \textbf{\textit{mean} Corruption Error (mCE)} is calculated as the average of the CE values across all corruption types:
\begin{equation}
    \text{mCE}^f = \frac{1}{|C|} \sum_{c \in C} \text{CE}_\text{c}^f
    \label{eq:mCE},
\end{equation}
with \( C \) representing the set of corruptions.
Moreover, for clean (In-Domain) samples, \textbf{Clean Error} is simply computed as
\(\left( E_{\text{clean}}^f \right) / \left( E_{\text{clean}}^{\text{baseline}} \right)\), where \( E_{\text{clean}}^f = 1 - \text{Acc}_{\text{clean}}^f \)
and \( \text{Acc}_{\text{clean}}^f \) is the Top-1 accuracy on the clean dataset.
This ensures that both the Clean Error and Corruption Error CE$_\text{c}$ are computed in a comparable manner, making it easy to assess the relative robustness of a model across both clean and corrupted settings along with Top-1 Average Accuracy\footnote{
    In addition to CE and \textit{m}CE, we also employ \textbf{Average Accuracy} as a performance measure and is defined as:
    \(
        \text{Avg. Acc}^f = \frac{1}{|C|} \sum_{c \in C} \left( \frac{1}{5} \sum_{s=1}^{5} \text{Acc}_{s,c}^f \right)
    \)
}, providing a comprehensive measure of degradation under out-of-distribution corruption scenarios.


\vspace{0.1cm}
\noindent\textbf{B. Implementation Setup:} In our experiments, $\mathbb{R}$MC's few-shot fine-tuning was performed on both Vision Transformer (ViT) and ResNet (RN) backbones. For each backbone, four configurations—using few-shot 1\%, 3\%, 7\%, and 10\% of the train set for tuning—were evaluated, resulting in a total of eight variants. All results are reported on the 10\% few-shot tuned $\mathbb{R}$MC, unless stated otherwise.
We initialize the $\mathbb{R}$MC-ViT and $\mathbb{R}$MC-RN models using BioMedCLIP and MedCLIP pretrained weights, respectively. We optimize the vision encoder using LoRA rank $r=16$ with the Adam optimizer across 20 epochs with a learning rate of $10^{-4}$ In computing the \textit{m}CE (as in Eq. \ref{eq:mCE}), the baseline model is set to OpenAI CLIP \cite{radford2021learning} consistently. Extended details are provided in \textbf{Appendix}.

\vspace{0.1cm}
\noindent\textbf{C. Comparative MVLMs:}
We compare $\mathbb{R}$MedCLIP against several state-of-the-art MVLMs. \textbf{OpenAI CLIP (2021)} \cite{radford2021learning} is pretrained on 400 million natural image–text pairs using standard augmentations, but not exposed to specific corruptions during training. \textbf{MedCLIP (2022)} \cite{wang2022medclipcontrastivelearningunpaired} fine-tunes on medical image–text pairs—typically chest X-rays \cite{johnson2019mimic} and retinal images—to improve diagnostic accuracy. \textbf{BioMedCLIP (2023)} \cite{zhang2025biomedclipmultimodalbiomedicalfoundation} further refines CLIP for medicine using the PMC-15M \cite{zhang2025biomedclipmultimodalbiomedicalfoundation} dataset---15M biomedical image-text pairs from PubMed Central \cite{pubmedcentral} archive---but omits corruption-based training. \textbf{UniMedCLIP (2025)} \cite{khattak2024unimedclipunifiedimagetextpretraining} trains on MedMNIST \cite{chen2021medmnist}, ROCO \cite{ruckert2024rocov2}, and PMC-OA \cite{lin2023pmc} datasets---ranging from multi-institutional imaging, radiology reports, to scientific articles---for cross-modal alignment but without any diversity exposure. These comparative MVLMs, though effective on clean data, as discussed in Section \ref{sec:results}, suffer notable performance degradation under out-of-distribution conditions.

{\renewcommand{\arraystretch}{1.2}
\begin{table}[H]
\caption{\small Clean Error, Corruption Error \textbf{CE}, and mean CE \textbf{mCE} comparison for ViT-B/16 backbone MVLMs across 
{\textcolor{orange}{Cell Microscopy}}, 
{\textcolor{myblue}{Breast Imaging}},
{\textcolor{teal}{Chest X-ray}}, 
{\textcolor{brown}{Fundoscopy}}, and 
{\textcolor{olive}{Retinal OCT}} modalities. 
Here, \textbf{Clean} denotes {Clean Error} on ``In-Distribution'' samples, while the rest denote OOD corruptions. The \textbf{mCE} is the mean CE across all corruptions (See Eq. \ref{eq:mCE}). \textbf{Bold} denotes best robustness while \underline{Underline} denotes second-best.
The Table that shows Accuracy metric is available in \textbf{Appendix}. 
}
\label{table:robustness}
\resizebox{\textwidth}{!}{%
\begin{tabular}{cllllllllll}
\multicolumn{1}{l}{\textbf{}} 
& \makecell{{\textcolor{orange}{Cell Microscopy~$\rightarrow$}}\\Methods~$\downarrow$}
 & \rotatebox{30}{\textbf{Clean}} 
 & \rotatebox{30}{Gauss.} 
 & \rotatebox{30}{Impulse} 
 & \rotatebox{30}{Motion} 
 & \rotatebox{30}{Zoom} 
 & \rotatebox{30}{Bright.} 
 & \rotatebox{30}{Contrast} 
 & \rotatebox{30}{Pixelate}  
 & \rotatebox{30}{\textbf{mCE}} \\ \hline
{}
 & CLIP        & \underline{100.0} & 100.0 & 100.0 & 100.0 & 100.0 & 100.0 & 100.0 & 100.0 & \underline{100.0} \\
 & MedCLIP     & 104.9 & 104.3 & 104.5 & 113.0 & 106.8 & 106.3 & 108.4 & 104.8 & 106.9 \\
 & BioMedCLIP  & 111.2 & 111.2 & 114.0 & 114.4 & 110.3 & 113.3 & 112.7 & 111.1 & 112.4 \\
 & UniMedCLIP  & 107.5 & 108.6 & 106.6 & 105.3 & 102.1 & 103.8 & 104.7 & 100.5 & 104.5 \\
 \rowcolor{skyblue!30} \cellcolor{white} \multirow{-5}{*}{\rotatebox{90}{\textcolor{purple}{\textbf{\scriptsize{MediMeta-C}}}}} & {$\mathbb{R}$MedCLIP}         & \textbf{24.3}  & 82.1  & 98.7  & 70.1  & 43.3  & 40.1  & 64.1  & 92.5  & \textbf{70.1}  \\ \hline
{}
 & CLIP        & 100.0 & 100.0 & 100.0 & 100.0 & 100.0 & 100.0 & 100.0 & 100.0 & 100.0 \\
 & MedCLIP     & 100.6 & 92.2  & 94.0  & 97.4  & 99.3  & 101.2 & 113.4 & 92.9  & 98.6  \\
 & BioMedCLIP  & \underline{86.9}  & 89.4  & 92.7  & 84.4  & 86.8  & 87.8  & 105.8 & 83.1  & \underline{90.0}  \\
 & UniMedCLIP  & 93.4  & 90.9  & 91.4  & 92.8  & 96.2  & 95.6  & 112.7 & 94.6  & 96.3  \\
 \rowcolor{skyblue!30} \cellcolor{white} \multirow{-5}{*}{\rotatebox{90}{\textcolor{blue}{\textbf{\scriptsize{MedMNIST-C}}}}} & {$\mathbb{R}$MedCLIP}         & \textbf{33.7}  & 53.3  & 54.9  & 42.1  & 51.3  & 40.7  & 64.4  & 48.2  & \textbf{50.7}  \\
\end{tabular}}

\resizebox{\textwidth}{!}{%
\begin{tabular}{cllllllllll}
\multicolumn{1}{l}{\textbf{}} 
& \makecell{{\textcolor{myblue}{Breast Img.~$\rightarrow$}}\\Methods~$\downarrow$}
 & \rotatebox{30}{\textbf{Clean}} 
 & \rotatebox{30}{Gauss.} 
 & \rotatebox{30}{Impulse} 
 & \rotatebox{30}{Motion} 
 & \rotatebox{30}{Zoom} 
 & \rotatebox{30}{Bright.} 
 & \rotatebox{30}{Contrast} 
 & \rotatebox{30}{Pixelate}  
 & \rotatebox{30}{\textbf{mCE}} \\ \hline
{}
 & OpenAI CLIP    & 100.0 & 100.0 & 100.0 & 100.0 & 100.0 & 100.0 & 100.0 & 100.0 & 100.0 \\
 & MedCLIP        & \underline{78.1}  & 79.9  & 78.6  & 67.3  & 70.0  & 75.9  & 74.3  & 72.4  & \underline{74.1}  \\
 & BioMedCLIP     & 90.1  & 71.4  & 64.0  & 99.7  & 92.3  & 95.3  & 96.2  & 103.5 & 88.9  \\
 & UniMedCLIP     & 102.6 & 107.5 & 101.1 & 100.7 & 100.5 & 100.4 & 112.6 & 110.1 & 104.7 \\
 \rowcolor{skyblue!30} \cellcolor{white} \multirow{-5}{*}{\rotatebox{90}{\textcolor{purple}{\textbf{\scriptsize{MediMeta-C}}}}} & {$\mathbb{R}$MedCLIP}         & \textbf{66.7}  & 70.0  & 66.2  & 65.3  & 65.4  & 65.4  & 74.9  & 70.2  & \textbf{68.2}  \\ \hline
{}
 & OpenAI CLIP    & 100.0 & 100.0 & 100.0 & 100.0 & 100.0 & 100.0 & 100.0 & 100.0 & 100.0 \\
 & MedCLIP        & 128.3 & 140.3 & 113.5 & 71.7  & 82.0  & 149.2 & 95.9  & 66.6  & 102.7 \\
 & BioMedCLIP     & 79.2  & 75.5  & 58.4  & 83.0  & 56.0  & 89.8  & 61.0  & 49.3  & \underline{67.6}  \\
 & UniMedCLIP     & \underline{79.2}  & 75.5  & 52.4  & 76.1  & 55.7  & 89.0  & 61.0  & 43.3  & \textbf{64.7}  \\
 \rowcolor{skyblue!30} \multirow{-5}{*}{\rotatebox{90}{\textcolor{blue}{\textbf{\scriptsize{MedMNIST-C}}}}} \cellcolor{white} & {$\mathbb{R}$MedCLIP}         & \textbf{75.5}  & 68.0  & 99.5  & 71.4  & 95.5  & 83.1  & 81.4  & 52.0  & 78.7  \\
\end{tabular}}


\resizebox{\textwidth}{!}{%
\begin{tabular}{cllllllllll}
\multicolumn{1}{l}{\textbf{}} 
& \makecell{{\textcolor{teal}{Chest X-ray~$\rightarrow$}}\\Methods~$\downarrow$}
 & \rotatebox{30}{\textbf{Clean}} 
 & \rotatebox{30}{Gauss.} 
 & \rotatebox{30}{Impulse} 
 & \rotatebox{30}{Motion} 
 & \rotatebox{30}{Zoom} 
 & \rotatebox{30}{Bright.} 
 & \rotatebox{30}{Contrast} 
 & \rotatebox{30}{Pixelate}  
 & \rotatebox{30}{\textbf{mCE}} \\ \hline
{}
 & OpenAI CLIP    & 100.0 & 100.0 & 100.0 & 100.0 & 100.0 & 100.0 & 100.0 & 100.0 & 100.0 \\
 & MedCLIP        & \underline{94.4}  & 104.7 & 95.5  & 100.5 & 94.6  & 94.7  & 96.0  & 95.4  & \underline{97.3}  \\
 & BioMedCLIP     & 100.0 & 100.7 & 94.0  & 100.0 & 99.8  & 99.9  & 99.0  & 95.6  & 98.4  \\
 & UniMedCLIP     & 100.0 & 100.7 & 94.3  & 97.3  & 93.8  & 99.9  & 101.6 & 94.6  & 97.5  \\
 \rowcolor{skyblue!30} \multirow{-5}{*}{\rotatebox{90}{\textcolor{purple}{\textbf{\scriptsize{MediMeta-C}}}}} \cellcolor{white} & {$\mathbb{R}$MedCLIP}         & \textbf{62.6}  & 91.8  & 86.0  & 65.5  & 58.7  & 74.7  & 84.4  & 84.9  & \textbf{78.0}  \\ \hline
{}
 & OpenAI CLIP    & 100.0 & 100.0 & 100.0 & 100.0 & 100.0 & 100.0 & 100.0 & 100.0 & 100.0 \\
 & MedCLIP        & \underline{45.4}  & 89.8  & 82.2  & 109.2 & 84.4  & 47.4  & 113.2  & 101.7   & \underline{89.7}  \\
 & BioMedCLIP     & 108.4 & 117.2 & 105.7 & 130.4 & 103.3 & 102.8 & 142.2 & 101.4 & 114.7 \\
 & UniMedCLIP     & 142.5 & 119.1 & 106.1 & 122.0 & 81.9  & 138.6 & 121.7 & 59.4  & 107.0 \\
 \rowcolor{skyblue!30} \multirow{-5}{*}{\rotatebox{90}{\textcolor{blue}{\textbf{\scriptsize{MedMNIST-C}}}}}  \cellcolor{white} & {$\mathbb{R}$MedCLIP}         & \textbf{29.7}  & 112.6  & 105.6  & 44.0  & 23.3  & 26.6  & 65.6  & 49.2  & \textbf{61.0}  \\
\end{tabular}}

\end{table}}

{\renewcommand{\arraystretch}{1.2}
\begin{table}[H]
\caption*{\small (Table Continued)}
\resizebox{\textwidth}{!}{%
\begin{tabular}{cllllllllll}
\multicolumn{1}{l}{\textbf{}} 
& \makecell{{\textcolor{brown}{Fundoscopy~$\rightarrow$}}\\Methods~$\downarrow$}
 & \rotatebox{30}{\textbf{Clean}} 
 & \rotatebox{30}{Gauss.} 
 & \rotatebox{30}{Impulse} 
 & \rotatebox{30}{Motion} 
 & \rotatebox{30}{Zoom} 
 & \rotatebox{30}{Bright.} 
 & \rotatebox{30}{Contrast} 
 & \rotatebox{30}{Pixelate}  
 & \rotatebox{30}{\textbf{mCE}} \\ \hline
{}
 & OpenAI CLIP    & 100.0 & 100.0 & 100.0 & 100.0 & 100.0 & 100.0 & 100.0 & 100.0 & \textbf{100.0} \\
 & MedCLIP        & 211.5 & 259.4 & 363.8 & 129.7 & 234.2 & 271.0 & 161.4    & 167.1  & 226.6 \\
 & BioMedCLIP     & 364.0 & 369.7 & 375.4 & 375.4 & 372.6 & 365.6 & 370.1 & 377.0 & 372.3 \\
 & UniMedCLIP     & \underline{97.1}  & 98.1  & 99.4  & 100.7 & 100.1 & 98.6  & 99.0  & 106.4 & \underline{100.3} \\
 \rowcolor{skyblue!30} \multirow{-5}{*}{\rotatebox{90}{\textcolor{purple}{\textbf{\scriptsize{MediMeta-C}}}}} \cellcolor{white} & {$\mathbb{R}$MedCLIP}         & \textbf{95.0}  & 109.9  & 253.6  & 100.7  & 124.9  & 96.8  & 130.5  & 122.1  & 134.1  \\ \hline
{}
 & OpenAI CLIP    & \textbf{100.0} & 100.0 & 100.0 & 100.0 & 100.0 & 100.0 & 100.0 & 100.0 & \textbf{100.0 }\\
 & MedCLIP        & \underline{101.8} & 146.1 & 127.5 & 141.0 & 127.4 & 113.2 & 88.8  & 114.6 & 122.7 \\
 & BioMedCLIP     & 118.8 & 135.0 & 129.7 & 120.8 & 121.6 & 127.6 & 98.8  & 122.3 & \underline{122.3} \\
 & UniMedCLIP     & 118.5 & 137.2 & 128.7 & 126.4 & 128.8 & 119.2 & 112.9 & 134.1 & 126.8 \\
 \rowcolor{skyblue!30} \multirow{-5}{*}{\rotatebox{90}{\textcolor{blue}{\textbf{\scriptsize{MedMNIST-C}}}}}  \cellcolor{white} & {$\mathbb{R}$MedCLIP}         & 126.4  & 143.5  & 138.2  & 135.6  & 139.1  & 122.5  & 114.1  & 142.7  & 133.7  \\
\end{tabular}}


\resizebox{\textwidth}{!}{%
\begin{tabular}{cllllllllll}
\multicolumn{1}{l}{\textbf{}} 
& \makecell{{\textcolor{olive}{Retinal OCT~$\rightarrow$}}\\Methods~$\downarrow$}
 & \rotatebox{30}{\textbf{Clean}} 
 & \rotatebox{30}{Gauss.} 
 & \rotatebox{30}{Impulse} 
 & \rotatebox{30}{Motion} 
 & \rotatebox{30}{Zoom} 
 & \rotatebox{30}{Bright.} 
 & \rotatebox{30}{Contrast} 
 & \rotatebox{30}{Pixelate}  
 & \rotatebox{30}{\textbf{mCE}} \\ \hline
{}
 & OpenAI CLIP    & 100.0 & 100.0 & 100.0 & 100.0 & 100.0 & 100.0 & 100.0 & 100.0 & 100.0 \\
 & MedCLIP        & \underline{83.6}  & 97.9 & 97.8 & 95.8 & 95.7 & 83.0 & 92.7 & 96.4  & \underline{94.2}  \\
 & BioMedCLIP     & 104.8 & 106.4 & 104.1 & 100.7 & 103.6 & 103.8 & 103.7 & 103.8 & 103.7 \\
 & UniMedCLIP     & 99.5  & 104.6 & 98.9  & 100.1 & 98.6  & 97.1  & 99.2  & 101.2 & 99.9  \\
 \rowcolor{skyblue!30} \multirow{-5}{*}{\rotatebox{90}{\textcolor{purple}{\textbf{\scriptsize{MediMeta-C}}}}} \cellcolor{white} & {$\mathbb{R}$MedCLIP}         & \textbf{42.7}  & 49.6  & 71.3  & 60.1  & 75.7  & 65.1  & 64.8  & 96.8  & \textbf{69.0}  \\ \hline
{}
 & OpenAI CLIP    & 100.0 & 100.0 & 100.0 & 100.0 & 100.0 & 100.0 & 100.0 & 100.0 & 100.0 \\
 & MedCLIP        & 114.8 & 102.9 & 108.0 & 131.5 & 120.1 & 112.3 & 105.4 & 103.1 & 111.9 \\
 & BioMedCLIP     & 100.5 & 96.3  & 100.5 & 102.1 & 98.4  & 100.2 & 94.7  & 96.6  & 98.4  \\
 & UniMedCLIP     & \underline{78.1}  & 76.3  & 91.5  & 91.2  & 91.4  & 73.9  & 94.0  & 90.5  & \underline{87.0}  \\
 \rowcolor{skyblue!30} \multirow{-5}{*}{\rotatebox{90}{\textcolor{blue}{\textbf{\scriptsize{MedMNIST-C}}}}}  \cellcolor{white} & {$\mathbb{R}$MedCLIP}         & \textbf{72.2}  & 67.9  & 83.2  & 48.9  & 76.7  & 46.1  & 54.3  & 92.6  & \textbf{67.1}  \\
\end{tabular}}
\vspace{-1cm}
\end{table}}

\begin{figure}
    \centering
    \includegraphics[width=1\linewidth]{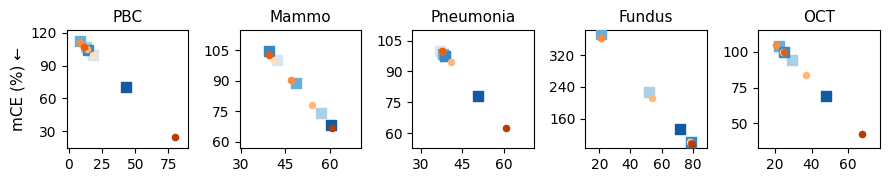}
    \includegraphics[width=1\linewidth]{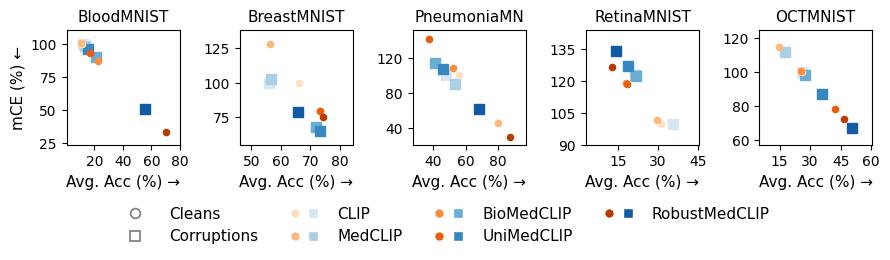}
    \vspace{-0.7cm}
    \caption{\small \textbf{Robustness vs. Accuracy trade-off} across modalities and MVLM baselines. Most MVLMs exhibit consistently high mCE and lower average accuracy across five modalities and two benchmarks.}
    \label{fig:mce_acc}
    \vspace{-0.4cm}
\end{figure}

\section{Results and Discussions}
\label{sec:results}
\subsection{Main Results}

\noindent\textbf{A. Robustness of MVLMs:}
The experimental results in Table \ref{table:robustness} reveal a variation in robustness across medical modalities. In Cell Microscopy, for example, although the baseline CLIP model consistently registers a clean error of 100\%, subsequent models such as MedCLIP, BioMedCLIP, and UniMedCLIP exhibit moderate error inflation under corruptions. Notably, our $\mathbb{R}$MC model shows a substantial reduction in clean error and a lower mCE, demonstrating that targeted robust adaptation can significantly mitigate the deleterious effects of visual distortions. In Breast Imaging, similar trends emerge; while the baseline and traditional MVLM variants suffer from pronounced error increases under corruption, our method consistently achieves lower mCE values, even when the absolute clean performance is slightly compromised. For modalities like Chest X-ray and Fundoscopy, the gap between clean and corrupted performance is less pronounced among models with higher intrinsic accuracy; however, the robustness of $\mathbb{R}$MC remains superior, highlighting that improvements in clean accuracy alone do not guarantee robustness. These observations indicate that the \textit{degradation observed under corrupted conditions is not simply a by-product of accuracy} enhancements. Rather, the ability of $\mathbb{R}$MC to leverage few-shot fine-tuning and robust (\textit{LoRA}) adaptation appears to directly counteract the effects of common image corruptions. 
\begin{HighlighterBox}{Not all MVLMs are consistently Robust but can certainly be!}
\small
    While many MVLMs excel on clean data, their performance can deteriorate markedly under realistic corruption scenarios—a challenge that $\mathbb{R}$MC's few-shot and low-rank tuning can help mitigate to some extent.
\end{HighlighterBox}

\vspace{0.1cm}
\noindent\textbf{B. MVLMs across Different Modalities:}
Although models such as MedCLIP, BioMedCLIP, and UniMedCLIP have been pretrained on various medical datasets, their performance across different modalities---Cell Microscopy, Fundoscopy, Breast Imaging, Chest X-ray, and Retinal OCT---varies considerably (Table \ref{table:robustness}). While baseline models maintain a clean error of 100\%, the relative increase in error under corruptions (as measured by mCE) is highly modality-dependent. For example, in the Cell Microscopy setting, our $\mathbb{R}$MC model achieves a dramatic reduction in clean error and mCE compared to other MVLMs, whereas in Breast Imaging, performance discrepancies are more pronounced. These results validate that the feature representations learned by current MVLMs are highly domain-specific, \textit{failing to generalize uniformly across clinical imaging tasks}. Improvements in clean accuracy do not always translate into robustness, as even models with competitive performance on clean images suffer under corruption (Fig. \ref{fig:mce_acc}). Thus, relying solely on MVLMs pretrained on \textit{few-modality} image–text pairs or on a single medical dataset is insufficient for clinical deployment. These observations underscore the critical need for diverse training data and robust adaptation strategies to achieve truly cross-modality generalization in MVLMs.
\begin{HighlighterBox}{Existing MVLMs are Not Modality Generalizable}
\small
    Existing MVLMs are inherently domain-specific; while few-shot robust adaptation enhances cross-modality robustness, large-scale training across diverse and numerous modalities is imperative for a truly generalizable MVLM.
\end{HighlighterBox}

\subsection{Discussions and Analyses}
\noindent\textbf{A. Trade-off between Accuracy and Robustness:}
Despite progressive improvements in average accuracy from MedCLIP to more recent UniMedCLIP, the observed tradeoff in Fig. \ref{fig:mce_acc} reveals that corruption robustness has not kept pace. While some MVLMs achieve higher average accuracy, their resilience under OOD corruptions remains limited, highlighting a gap between clean-data success and robust generalization. Notably, MedCLIP maintains consistent results in one modality yet falters in others, reflecting a lack of universal cross-modal adaptability unlike $\mathbb{R}$MC. This discrepancy suggests that simply enhancing average accuracy does not guarantee improved corruption resistance. Instead, specialized strategies—such as domain-aware fine-tuning—appear essential for bridging the accuracy–robustness divide across diverse clinical imaging modalities. Robust adaptation remains essential for deployment.

\begin{figure}[t]
    \centering
    \includegraphics[width=1\linewidth]{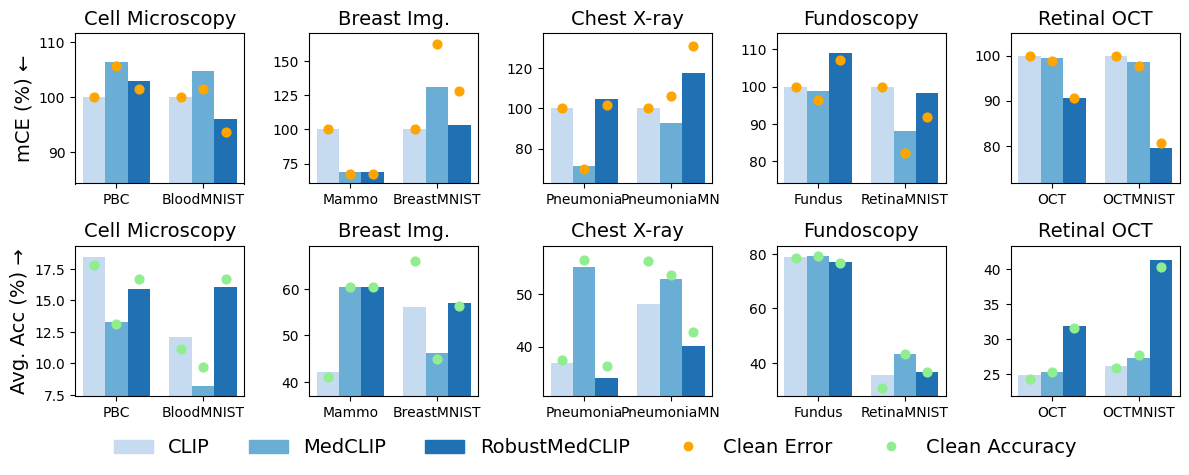}
    \vspace{-0.7cm}
    \caption{\small mCE and Accuracy comparison of \textbf{ResNet-50-based MVLMs} (MedCLIP and our $\mathbb{R}$MC) against the CLIP baseline across MediMeta-C and MedMNIST-C benchmarks. \textbf{MN} indicates the abbreviation for MNIST where applicable.}
    \label{fig:resnet_backbone}
    \vspace{-0.4cm}
\end{figure}

\vspace{0.1cm}
\noindent\textbf{B. Impact of Backbones on Robustness:}
ResNet-based MVLMs in Fig. \ref{fig:resnet_backbone} show limited gains in robustness compared to their ViT counterparts, as evidenced by the similar mCE bars across CLIP, MedCLIP, and $\mathbb{R}$MC in multiple modalities. While MedCLIP occasionally reduces mCE slightly, it does not outperform CLIP consistently, suggesting that ResNet architectures alone do not guarantee stronger corruption resistance. In the lower row, MedCLIP likewise fails to achieve substantially higher accuracy than CLIP, indicating that improved backbone capacity does not automatically translate to enhanced performance. Notably, our $\mathbb{R}$MC approach yields a higher average accuracy, surpassing both CLIP and MedCLIP in several modalities, yet still reflects only moderate gains in mCE. Overall, ResNet-based models remain less robust than ViT-based approaches.

\begin{figure}[t]
    \centering
    \includegraphics[width=1\linewidth]{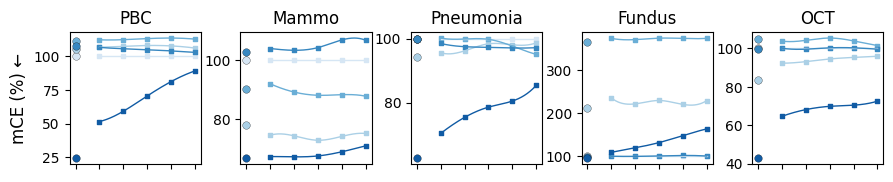}
    \includegraphics[width=1\linewidth]{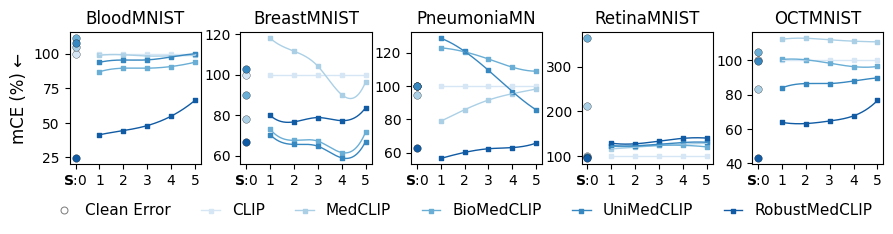}
    \vspace{-0.7cm}
    \caption{\small Performance Degradation of Medical VLMs Across Five Corruption Severity Levels in terms of mCE. \textbf{S} means Severity Level while \textbf{S}:0 implies Clean Error.}
    \label{fig:severity_levels}
    \vspace{-0.3cm}
\end{figure}

\vspace{0.1cm}
\noindent\textbf{C. Performance Against Severity Levels:}
Prior MVLMs exhibit systemic fragility under escalating corruption severity, with mCE increasing on average as distortions intensify (Fig. \ref{fig:severity_levels}). These MVLMs struggle to retain discriminative features under artifacts. While having modality-agnostic design and \textit{supposedly} broadly applicable, they fail to prioritize corruption-invariant features, leading to erratic performance in modalities prone to specific distortions (e.g., motion blur in OCT).
Notably, the impact of severity varies by imaging modality---in Fundoscopy for instance---all models display relative resilience, suggesting anatomical context mitigates distortion effects. Our $\mathbb{R}$MC achieves superior robustness on clean and mildly corrupted samples, attributed to few-shot adaptation that preserves feature integrity. At higher severities, $\mathbb{R}$MC remains competitive, leveraging adapted few-shot representations to counter progressive degradation. 

\begin{figure}[t]
    \centering
    \includegraphics[width=1.0\linewidth]{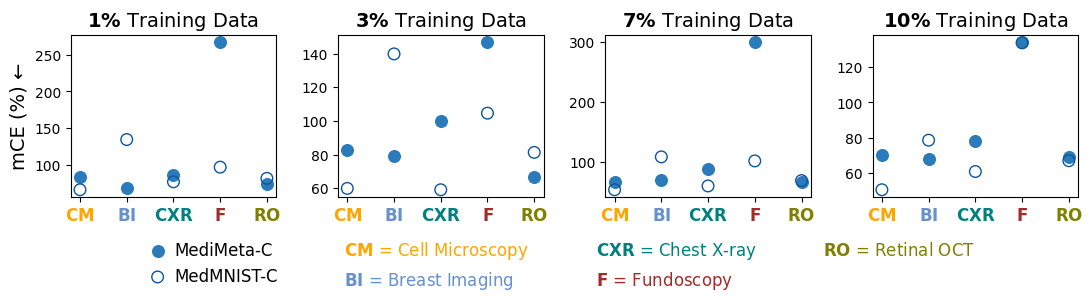}
    \vspace{-0.8cm}
    \caption{\small \textbf{Effect of Few-shot Samples} on Fine-Tuning $\mathbb{R}$MC. Performance of $\mathbb{R}$MC across five modalities with varying percentages of clean training data. $\mathbb{R}$MC model.}
    \label{fig:few_shot_ablation}
    \vspace{-0.4cm}
\end{figure}

\vspace{0.1cm}
\noindent\textbf{D. Ablation of Few-Shot Samplings:}
Fig. \ref{fig:few_shot_ablation} illustrates the few-shot performance of $\mathbb{R}$MC across five medical imaging modalities with varying proportions of clean training data. While performance generally improves with increased data, gains are highly modality-dependent rather than strictly linear. Notably, modalities such as Fundoscopy maintain stable robustness regardless of data volume, suggesting insensitivity to sample size. In contrast, Chest X-ray and Retinal OCT exhibit sharp mCE reductions (e.g., ~18\% drop from 1\% to 10\% data), highlighting greater data efficiency. Breast Imaging and Cell Microscopy show more gradual improvements, likely due to intrinsic noise or task complexity. These results confirm that few-shot adaptation effectively mitigates corruption-induced degradation without compromising generalization, especially in resource-constrained clinical scenarios.

\begin{HighlighterBox}{Discussions at a Glance}
\small
\begin{enumerate}[label=\Alph*., leftmargin=*]
    \item Increasing accuracy alone does not ensure robustness, especially under corruptions and across modalities.
    \item Backbone choice influences performance, but ResNet-based MVLMs lag behind ViT in robustness.
    \item Corruption severity critically degrades MVLMs, with modality-specific vulnerabilities emerging.
    \item Few-shot fine-tuning improves robustness efficiently, especially in data-limited medical settings.
\end{enumerate}
\end{HighlighterBox}

\vspace{-0.15cm}
\section{Conclusion}
\vspace{-0.15cm}
\noindent\textbf{A. Summary:} Our study demonstrates that enhancing robustness in Medical Vision-Language Models (MVLMs) requires a paradigm shift from maximizing clean accuracy to optimizing for resilience under distribution shifts. Our extensive evaluations, conducted using the comprehensive MediMeta-C and MedMNIST-C benchmarks, reveal that while baseline models such as MedCLIP, BioMedCLIP, and UniMedCLIP achieve high accuracy on pristine datasets, they suffer significant increases in mean Corruption Error (mCE) when exposed to realistic corruptions. In contrast, our proposed $\mathbb{R}$obustMedCLIP, which leverages few-shot fine-tuning with low-rank adaptation on a diverse set of clinical domains, achieves markedly lower clean errors and mCE. These results underscore that \textit{data-modality diversity is paramount over dataset volume} for achieving robust cross-modality generalization. The analysis further indicates that variations in backbone architecture and corruption severity yield modality-specific performance gaps, highlighting the need for tailored adaptation strategies rather than relying solely on improvements in clean accuracy.

\vspace{0.1cm}
\noindent\textbf{B. Future Directions:} Future work should investigate integrating adaptive, parameter-efficient tuning mechanisms across a wider spectrum of OOD clinical domains to further enhance robustness. Expanding the evaluation framework to encompass additional corruption types and real-world clinical data will be crucial in developing MVLMs that are both accurate and robust in deployment.

{
\bibliographystyle{splncs04}
\bibliography{refs}
}

\newpage
\appendix
\section*{\begin{huge} \textbf{Appendix} \end{huge}}
\label{appendix:main}

{\renewcommand{\arraystretch}{1.2}
\begin{table}[H]
\caption{\small \textbf{Clean} Accuracy, Accuracy against Corruptions, and Average Accuracy (\textbf{A.Acc.}) comparison for ViT-B/16 backbone MVLMs across 
{\textcolor{orange}{Cell Microscopy}}, 
{\textcolor{myblue}{Breast Imaging}},
{\textcolor{teal}{Chest X-ray}}, 
{\textcolor{brown}{Fundoscopy}}, and 
{\textcolor{olive}{Retinal OCT}} modalities. 
Here, \textbf{Clean} denotes Top-1 Accuracy on clean ``In-Distribution'' samples, while \textbf{A.Acc.} is the average Accuracy across all corruptions (See Eq. \ref{eq:acc} in main paper). \textbf{Bold} denotes best accuracy while \underline{Underline} denotes second-best.
}
\label{table:accuracy}
\resizebox{\textwidth}{!}{%
\begin{tabular}{cllllllllll}
\multicolumn{1}{l}{\textbf{}} 
& \makecell{{\textcolor{orange}{Cell Microscopy~$\rightarrow$}}\\Methods~$\downarrow$}
 & \rotatebox{30}{\textbf{Clean}} 
 & \rotatebox{30}{Gauss.} 
 & \rotatebox{30}{Impulse} 
 & \rotatebox{30}{Motion} 
 & \rotatebox{30}{Zoom} 
 & \rotatebox{30}{Bright.} 
 & \rotatebox{30}{Contrast} 
 & \rotatebox{30}{Pixelate}  
 & \rotatebox{30}{\textbf{A.Acc.}} \\ \hline
{}
 & CLIP        & \underline{17.81} & 17.58 & 17.96 & 20.09 & 17.20 & 19.30 & 18.84 & 17.94 & \underline{18.42} \\
 & MedCLIP     & 13.75 & 14.02 & 14.25 & 9.68  & 11.54 & 14.26 & 12.02 & 13.99 & 12.82 \\
 & BioMedCLIP  & 8.60  & 8.32  & 6.49  & 8.60  & 8.68  & 8.55  & 8.53  & 8.80  & 8.28  \\
 & UniMedCLIP  & 11.61 & 10.46 & 12.56 & 15.89 & 15.45 & 16.23 & 14.99 & 17.56 & 14.74 \\
 \rowcolor{skyblue!30} \multirow{-5}{*}{\rotatebox{90}{\textcolor{purple}{\textbf{\scriptsize{MediMeta-C}}}}} \cellcolor{white}  & {$\mathbb{R}$MedCLIP} & \textbf{80.05} & 32.30 & 19.06 & 43.95 & 64.11 & 67.60 & 47.98 & 24.10 & \textbf{42.73} \\ \hline
{}
 & CLIP        & 11.11 & 8.65  & 9.12  & 9.98  & 9.83  & 13.31 & 23.41 & 10.38 & 12.10 \\
 & MedCLIP     & 10.55 & 15.78 & 14.55 & 12.30 & 10.47 & 12.24 & 13.13 & 16.78 & 13.61 \\
 & BioMedCLIP  & \underline{22.71} & 18.35 & 15.74 & 24.03 & 21.74 & 23.89 & 18.98 & 25.48 & \underline{21.17} \\
 & UniMedCLIP  & 16.98 & 16.92 & 16.93 & 16.42 & 13.29 & 17.15 & 13.69 & 15.21 & 15.66 \\
 \rowcolor{skyblue!30} \multirow{-5}{*}{\rotatebox{90}{\textcolor{blue}{\textbf{\scriptsize{MedMNIST-C}}}}} \cellcolor{white} & {$\mathbb{R}$MedCLIP} & \textbf{70.07} & 51.29 & 50.11 & 62.09 & 53.78 & 64.69 & 50.69 & 56.81 & \textbf{55.64} \\
\end{tabular}}

\resizebox{\textwidth}{!}{%
\begin{tabular}{cllllllllll}
\multicolumn{1}{l}{\textbf{}} 
& \makecell{{\textcolor{myblue}{Breast Img.~$\rightarrow$}}\\Methods~$\downarrow$}
 & \rotatebox{30}{\textbf{Clean}} 
 & \rotatebox{30}{Gauss.} 
 & \rotatebox{30}{Impulse} 
 & \rotatebox{30}{Motion} 
 & \rotatebox{30}{Zoom} 
 & \rotatebox{30}{Bright.} 
 & \rotatebox{30}{Contrast} 
 & \rotatebox{30}{Pixelate}  
 & \rotatebox{30}{\textbf{A.Acc.}} \\ \hline
{}
 & CLIP        & 41.10 & 43.80 & 40.25 & 40.00 & 39.88 & 39.82 & 46.32 & 45.09 & 42.16 \\
 & MedCLIP     & 53.99 & 55.09 & 53.01 & 59.63 & 57.91 & 54.29 & 60.12 & 60.25 & 57.19 \\
 & BioMedCLIP  & \underline{46.93} & 59.88 & 61.78 & 40.18 & 44.48 & 42.64 & 48.34 & 43.19 & \underline{48.64} \\
 & UniMedCLIP  & 39.57 & 39.57 & 39.57 & 39.57 & 39.57 & 39.57 & 39.57 & 39.57 & 39.57 \\
 \rowcolor{skyblue!30} \multirow{-5}{*}{\rotatebox{90}{\textcolor{purple}{\textbf{\scriptsize{MediMeta-C}}}}} \cellcolor{white} & {$\mathbb{R}$MedCLIP} & \textbf{60.74} & 60.67 & 60.43 & 60.80 & 60.67 & 60.61 & 59.82 & 61.47 & \textbf{60.64} \\ \hline
{}
 & CLIP        & 66.03 & 64.36 & 48.59 & 64.62 & 51.67 & 69.74 & 55.90 & 37.82 & 56.10 \\
 & MedCLIP     & 56.41 & 50.00 & 41.67 & 74.62 & 60.38 & 54.87 & 57.69 & 58.59 & 56.83 \\
 & BioMedCLIP  & 73.08 & 73.08 & 70.00 & 70.64 & 72.95 & 72.82 & 73.08 & 69.36 & \underline{71.70} \\
 & UniMedCLIP  & \underline{73.08} & 73.08 & 73.08 & 73.08 & 73.08 & 73.08 & 73.08 & 73.08 & \textbf{73.08} \\
 \rowcolor{skyblue!30} \multirow{-5}{*}{\rotatebox{90}{\textcolor{blue}{\textbf{\scriptsize{MedMNIST-C}}}}} \cellcolor{white} & {$\mathbb{R}$MedCLIP} & \textbf{74.36} & 75.77 & 48.85 & 74.74 & 53.85 & 74.87 & 64.10 & 67.69 & 65.70 \\
\end{tabular}}


\resizebox{\textwidth}{!}{%
\begin{tabular}{cllllllllll}
\multicolumn{1}{l}{\textbf{}} 
& \makecell{{\textcolor{teal}{Chest X-ray~$\rightarrow$}}\\Methods~$\downarrow$}
 & \rotatebox{30}{\textbf{Clean}} 
 & \rotatebox{30}{Gauss.} 
 & \rotatebox{30}{Impulse} 
 & \rotatebox{30}{Motion} 
 & \rotatebox{30}{Zoom} 
 & \rotatebox{30}{Bright.} 
 & \rotatebox{30}{Contrast} 
 & \rotatebox{30}{Pixelate}  
 & \rotatebox{30}{\textbf{A.Acc.}} \\ \hline
{}
 & CLIP        & 37.50 & 37.95 & 33.75 & 37.50 & 37.40 & 37.47 & 37.66 & 37.02 & 36.96 \\
 & MedCLIP     & \underline{41.03} & 35.03 & 36.73 & 37.18 & 40.77 & 40.80 & 40.16 & 39.90 & \underline{38.65} \\
 & BioMedCLIP  & 37.50 & 37.50 & 37.72 & 37.50 & 37.50 & 37.50 & 38.27 & 39.81 & 37.97 \\
 & UniMedCLIP  & 37.50 & 37.50 & 37.50 & 39.17 & 41.31 & 37.50 & 36.67 & 40.45 & 38.59 \\
 \rowcolor{skyblue!30} \multirow{-5}{*}{\rotatebox{90}{\textcolor{purple}{\textbf{\scriptsize{MediMeta-C}}}}} \cellcolor{white} & {$\mathbb{R}$MedCLIP} & \textbf{60.90} & 43.01 & 43.01 & 59.07 & 63.24 & 53.27 & 47.40 & 46.54 & \textbf{50.79} \\ \hline
{}
 & CLIP        & 56.25 & 47.47 & 41.09 & 52.85 & 41.15 & 55.10 & 59.62 & 38.91 & 48.03 \\
 & MedCLIP     & \underline{80.13} & 52.82 & 51.57 & 48.49 & 50.32 & 78.72 & 54.29 & 37.88 & \underline{53.44} \\
 & BioMedCLIP  & 52.56 & 38.43 & 37.72 & 38.53 & 39.20 & 53.85 & 42.56 & 38.04 & 41.19 \\
 & UniMedCLIP  & 37.66 & 37.44 & 37.50 & 42.50 & 51.83 & 37.76 & 50.83 & 63.69 & 45.93 \\
 \rowcolor{skyblue!30} \multirow{-5}{*}{\rotatebox{90}{\textcolor{blue}{\textbf{\scriptsize{MedMNIST-C}}}}} \cellcolor{white} & {$\mathbb{R}$MedCLIP} & \textbf{87.02} & 40.83 & 37.79 & 79.26 & 86.31 & 88.08 & 73.49 & 69.97 & \textbf{67.96} \\
\end{tabular}}

\end{table}}

{\renewcommand{\arraystretch}{1.2}
\begin{table}[t]
\caption*{\small (Table Continued)}
\resizebox{\textwidth}{!}{%
\begin{tabular}{cllllllllll}
\multicolumn{1}{l}{\textbf{}} 
& \makecell{{\textcolor{brown}{Fundoscopy~$\rightarrow$}}\\Methods~$\downarrow$}
 & \rotatebox{30}{\textbf{Clean}} 
 & \rotatebox{30}{Gauss.} 
 & \rotatebox{30}{Impulse} 
 & \rotatebox{30}{Motion} 
 & \rotatebox{30}{Zoom} 
 & \rotatebox{30}{Bright.} 
 & \rotatebox{30}{Contrast} 
 & \rotatebox{30}{Pixelate}  
 & \rotatebox{30}{\textbf{A.Acc.}} \\ \hline
{}
 & CLIP        & \underline{78.28} & 78.62 & 78.94 & 78.94 & 78.78 & 78.38 & 78.81 & 79.03 & \textbf{78.79} \\
 & MedCLIP     & 54.06 & 44.56 & 23.38 & 72.69 & 50.31 & 41.41 & 65.81 & 64.97 & 51.88 \\
 & BioMedCLIP  & 20.94 & 20.97 & 20.94 & 20.94 & 20.94 & 20.94 & 21.59 & 20.94 & 21.04 \\
 & UniMedCLIP  & 78.91 & 79.03 & 79.06 & 78.78 & 78.75 & 78.69 & 79.03 & 77.69 & \underline{78.72} \\
 \rowcolor{skyblue!30} \multirow{-5}{*}{\rotatebox{90}{\textcolor{purple}{\textbf{\scriptsize{MediMeta-C}}}}} \cellcolor{white} & {$\mathbb{R}$MedCLIP} & \textbf{79.38} & 76.50 & 46.59 & 78.78 & 73.50 & 79.06 & 72.34 & 74.41 & 71.60 \\ \hline
{}
 & CLIP        & \textbf{31.00} & 39.50 & 35.95 & 34.75 & 36.35 & 36.10 & 28.00 & 38.10 & \textbf{35.54} \\
 & MedCLIP     & \underline{29.75} & 11.60 & 18.35 &  8.00 & 18.90 & 27.65 & 36.10 & 29.05 & 21.38 \\
 & BioMedCLIP  & 18.00 & 18.30 & 16.90 & 21.20 & 22.60 & 18.45 & 28.85 & 24.30 & \underline{21.51} \\
 & UniMedCLIP  & 18.25 & 17.00 & 17.55 & 17.55 & 18.00 & 23.85 & 18.70 & 17.00 & 18.52 \\
 \rowcolor{skyblue!30} \multirow{-5}{*}{\rotatebox{90}{\textcolor{blue}{\textbf{\scriptsize{MedMNIST-C}}}}} \cellcolor{white} & {$\mathbb{R}$MedCLIP} & 12.75 & 13.20 & 11.50 & 11.50 & 11.45 & 21.70 & 17.85 & 11.65 & 14.12 \\
\end{tabular}}


\resizebox{\textwidth}{!}{%
\begin{tabular}{cllllllllll}
\multicolumn{1}{l}{\textbf{}} 
& \makecell{{\textcolor{olive}{Retinal OCT~$\rightarrow$}}\\Methods~$\downarrow$}
 & \rotatebox{30}{\textbf{Clean}} 
 & \rotatebox{30}{Gauss.} 
 & \rotatebox{30}{Impulse} 
 & \rotatebox{30}{Motion} 
 & \rotatebox{30}{Zoom} 
 & \rotatebox{30}{Bright.} 
 & \rotatebox{30}{Contrast} 
 & \rotatebox{30}{Pixelate}  
 & \rotatebox{30}{\textbf{A.Acc.}} \\ \hline
{}
 & CLIP        & 24.30 & 26.26 & 24.98 & 24.84 & 24.58 & 23.02 & 24.70 & 25.94 & 24.90 \\
 & MedCLIP     & \underline{36.70} & 27.84 & 26.60 & 27.98 & 27.80 & 36.08 & 30.16 & 28.60 & \underline{29.29} \\
 & BioMedCLIP  & 20.70 & 21.52 & 21.90 & 24.34 & 21.88 & 20.10 & 21.90 & 23.16 & 22.11 \\
 & UniMedCLIP  & 24.70 & 22.84 & 25.84 & 24.76 & 25.66 & 25.24 & 25.34 & 25.08 & 24.97 \\
 \rowcolor{skyblue!30} \multirow{-5}{*}{\rotatebox{90}{\textcolor{purple}{\textbf{\scriptsize{MediMeta-C}}}}} \cellcolor{white} & {$\mathbb{R}$MedCLIP} & \textbf{67.70} & 63.46 & 46.48 & 54.82 & 42.92 & 49.92 & 51.20 & 28.34 & \textbf{48.16} \\ \hline
{}
 & CLIP        & 25.90 & 22.72 & 27.24 & 28.26 & 28.52 & 25.98 & 25.74 & 25.02 & 26.21 \\
 & MedCLIP     & 14.90 & 20.48 & 21.40 &  5.68 & 14.18 & 16.86 & 21.76 & 22.68 & 17.58 \\
 & BioMedCLIP  & 25.50 & 25.58 & 26.88 & 26.74 & 29.64 & 25.82 & 29.64 & 27.54 & 27.41 \\
 & UniMedCLIP  & \underline{42.10} & 41.06 & 33.42 & 34.58 & 34.68 & 45.32 & 30.16 & 32.16 & \underline{35.91} \\
 \rowcolor{skyblue!30} \multirow{-5}{*}{\rotatebox{90}{\textcolor{blue}{\textbf{\scriptsize{MedMNIST-C}}}}} \cellcolor{white} & {$\mathbb{R}$MedCLIP} & \textbf{46.50} & 47.52 & 39.44 & 64.94 & 45.20 & 65.90 & 59.68 & 30.54 & \textbf{50.46} \\
\end{tabular}}
\vspace{-0.15cm}
\end{table}}

\noindent\textbf{A. Additional Implementation Details:} 
For fine-tuning, $\mathbb{R}$MC-ViT is initialized with pretrained weights from BioMedCLIP with ViT-B/16, while $\mathbb{R}$MC-ResNet (RN50) uses the MedCLIP RN50 variant. Few-shot tuning is performed using LoRA with a rank of $r=16$, optimized with the Adam optimizer for 20 epochs at a learning rate of $10^{-4}$. 
To evaluate robustness, the \textit{mean} Corruption Error (\textit{m}CE) is computed using OpenAI CLIP~\cite{radford2021learning} with a ViT-B/16 backbone as the corruption robustness baseline, as defined in Eq.~\ref{eq:mCE}. 
This choice of baseline provides a standardized and model-agnostic point of reference, allowing for consistent comparisons of corruption robustness across both ViT and RN50-based MVLMs, including $\mathbb{R}$MC.

\vspace{0.1cm}
\noindent\textbf{B. Computation and Parameter Scaling of $\mathbb{R}$MC Variants:}
Table~\ref{tab:performance_metrics} presents parameter analysis of $\mathbb{R}$MC variants trained on ViT and ResNet backbones. Notably, $\mathbb{R}$MC-ViT achieves strong performance with only 1.02\% of parameters fine-tuned via LoRA, maintaining competitive accuracy even with minimal sampling (e.g., 66.45\% at 3\% samples). As sample size increases, ViT shows marked gains, peaking at 80.05\% accuracy with only a modest training time of 1.62 hours. In contrast, $\mathbb{R}$MC-RN exhibits limited accuracy gains despite higher parameter exposure (1.39\%) and similar runtime, suggesting underutilization of representational capacity. Furthermore, $\mathbb{R}$MC-ViT consistently outperforms its ResNet counterpart in both clean and corrupted settings, all while maintaining a stable computational footprint. This indicates that transformer-based MVLMs offer a more favorable robustness-efficiency trade-off, especially under few-shot constraints. 

{\renewcommand{\arraystretch}{1.5}
\begin{table}[t]
\caption{\small Dataset statistics for MediMeta~\cite{woerner2024comprehensive} across five imaging modalities.}
\centering
\resizebox{\textwidth}{!}{%
\begin{tabular}{llllll}
\textbf{Modality~$\downarrow$} & \textbf{Data Name} & \textbf{\#Train/Val/Test}~~ & \textbf{\#Classes}~~~~~~~~~ & \textbf{Description}~~~~ & \textbf{Class Labels} \\ \hline
\textcolor{orange}{Cell Microscopy} 
    & PBC & 11964/1709/3149 & Multi-Class (8) & Blood cells & 
    \makecell[l]{basophil, eosinophil, erythroblast,\\immature granulocyte, lymphocyte,\\monocyte, neutrophil, platelet} \\
\textcolor{myblue}{Breast Imaging} 
    & Mammo & 1332/214/326 & Binary (2) & Calcifications & 
    \makecell[l]{malignant, benign} \\
\textcolor{teal}{Chest X-ray} 
    & Pneumonia & 4415/817/624 & Multi-Class (3) & Lung infection & 
    \makecell[l]{normal, bacteria, virus} \\
\textcolor{brown}{Fundoscopy} 
    & Fundus & 1920/640/640 & Binary (2) & Eye diseases & 
    \makecell[l]{abnormal, normal} \\
\textcolor{olive}{Retinal OCT} 
    & OCT & 91615/16694/1000 & Multi-Class (4) & Retinal layers & 
    \makecell[l]{cnv, normal, dme, drusen} \\
\end{tabular}}
\label{table:medimeta_data}
\vspace{-0.3cm}
\end{table}}

{\renewcommand{\arraystretch}{1.5}
\begin{table}[t]
\caption{\small Dataset statistics for MedMNIST~\cite{chen2021medmnist} across five imaging modalities.}
\centering
\resizebox{\textwidth}{!}{%
\begin{tabular}{llllll}
\textbf{Modality~$\downarrow$} & \textbf{Data Name} & \textbf{\#Train/Val/Test}~~ & \textbf{\#Classes}~~~~~~~~~ & \textbf{Description}~~~~ & \textbf{Class Labels} \\ \hline
\textcolor{orange}{Cell Microscopy} 
    & BloodMNIST & 11959/1712/3421 & Multi-Class (8) & Blood cells & 
    \makecell[l]{basophil, eosinophil, erythroblast,\\granulocytes, lymphocyte, monocyte,\\neutrophil, platelet} \\
\textcolor{myblue}{Breast Imaging} 
    & BreastMNIST & 546/78/156 & Binary (2) & Breast tumors & 
    \makecell[l]{malignant, benign} \\
\textcolor{teal}{Chest X-ray} 
    & PneumoniaMNIST & 4708/524/624 & Binary (2) & Lung infection & 
    \makecell[l]{normal, pneumonia} \\
\textcolor{brown}{Fundoscopy} 
    & RetinaMNIST & 1080/120/400 & Multi-Class (5) & Eye diseases & 
    \makecell[l]{0, 1, 2, 3, 4} \\
\textcolor{olive}{Retinal OCT} 
    & OCTMNIST & 97477/10832/1000 & Multi-Class (4) & Retinal layers & 
    \makecell[l]{choroidal neovascularization, \\diabetic macular edema,\\drusen, normal} \\
\end{tabular}}
\label{table:medmnist_data}
\vspace{-0.5cm}
\end{table}}

{\renewcommand{\arraystretch}{1.2}
\begin{table}[]
\centering
\caption{Performance comparison of few-shot $\mathbb{R}$MC-ViT and $\mathbb{R}$MC-ResNet variants on \textbf{clean} (MediMeta) and \textbf{corrupted} (MediMeta-C) datasets. Computation metrics include Training Time (in hours), Total Parameters, and the percentage of Trainable Parameters. Here, \textbf{M} denotes parameters in millions.}
\vspace{0.2cm}
\label{tab:performance_metrics}
\resizebox{\textwidth}{!}{%
\begin{tabular}{lcccccccc}
\multicolumn{2}{l}{{\textcolor{orange}{Cell Microscopy}}}
& \multicolumn{2}{c}{\makecell[c]{\textcolor{purple}{MediMeta}\\\textbf{Cleans}}} 
& \multicolumn{2}{c}{\makecell[c]{\textcolor{purple}{MediMeta-C}\\\textbf{Corruptions}}} 
& \multicolumn{3}{c}{\textbf{Computational Statistics}} \\
\cmidrule(lr){3-4} \cmidrule(r){5-6} \cmidrule(){7-9}

\makecell{$\mathbb{R}$MC\\Variant}~~~~~~ 
& \makecell[c]{Few-Shots\\\%}~ 
& \makecell{Avg.Acc.\\$\uparrow$}~~ & \makecell{Error\\$\downarrow$}~~~ & \makecell{Avg.Acc.\\$\uparrow$}~~ & \makecell{mCE\\$\downarrow$}~~~~  & \makecell[l]{Train\\Time(hrs)} & \makecell[l]{Total\\Params}~~ & \makecell[l]{Trainable\\Params (\%)} \\
\hline
\multirow{4}{*}{$\mathbb{R}$MC-ViT} 
    & 1  & 46.36 & 65.3  & 31.23 & 84.3  & 0.43 & 87M & 1.02 \\
    & 3  & 66.45 & 40.8  & 32.23 & 83.0  & 0.68 & 87M & 1.02 \\
    & 7  & 73.30 & 32.5  & \textbf{44.83} & \textbf{67.5}  & 1.28 & 87M & 1.02 \\
\rowcolor{skyblue!30} \cellcolor{white} & 10 & \textbf{80.05} & \textbf{24.3}  & 42.73 & 70.1  & 1.62 & 87M & 1.02 \\
\hline
\multirow{4}{*}{$\mathbb{R}$MC-ResNet} 
    & 1  & 12.84 & 106   & 12.92 & 106.7 & 0.22 & 49M & 1.39 \\
    & 3  & 12.81 & 103.1 & 12.35 & 107.4 & 0.33 & 49M & 1.39 \\
    & 7  & 14.77 & 103.7 & 13.50 & 106.0 & 0.61 & 49M & 1.39 \\
\rowcolor{skyblue!30} \cellcolor{white} & 10 & \textbf{16.67} & \textbf{101.4} & \textbf{15.93} & \textbf{103.0} & 0.80 & 49M & 1.39 \\
\vspace{-1cm}
\end{tabular}}
\end{table}}

\begin{figure}[t]
    \centering
    \includegraphics[width=1.0\linewidth]{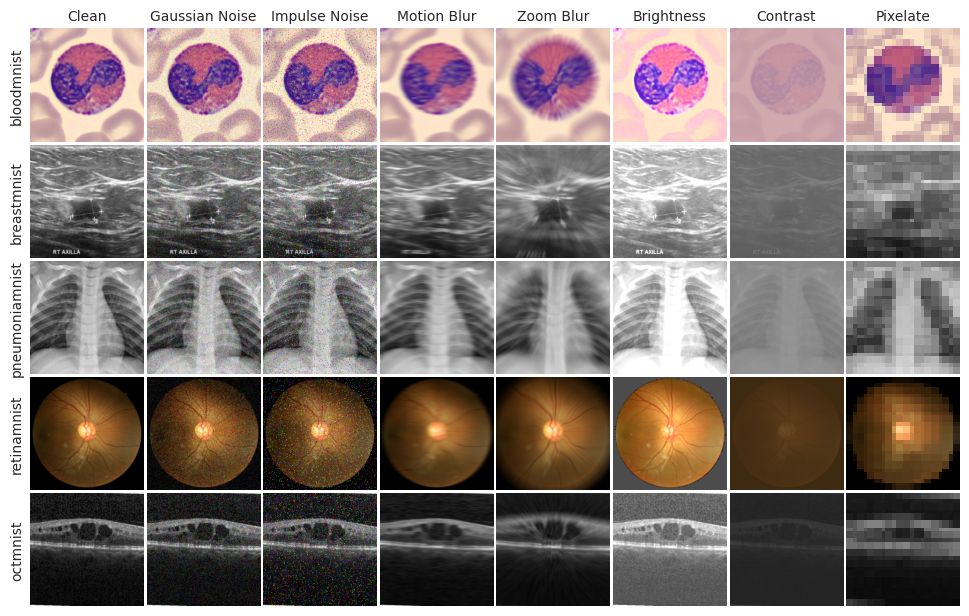}
    \vspace{-0.8cm}
    \caption{\small \textbf{Corrupted samples} from MedMNIST-C \cite{di2024medmnist} dataset. The y-axis shows dataset names by modality and the x-axis displays corruption types at a fixed severity level.}
    \label{fig:medmnist_c}
    \vspace{-0.5cm}
\end{figure}

\begin{figure*}[h]
    \centering
    \includegraphics[width=1.0\textwidth]{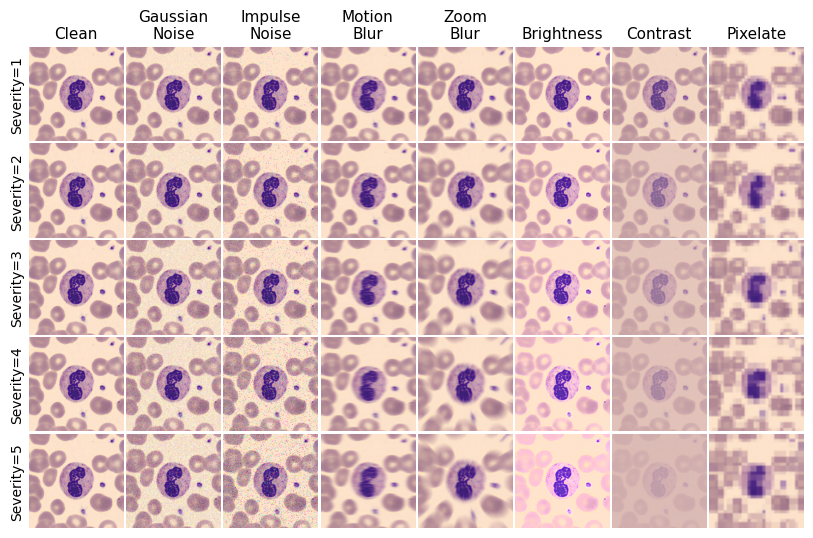}
    \vspace{-0.8cm}
    \caption{\small Example images from cell microscopy modality of MediMeta-C -- \textbf{PBC-C}, illustrating corruptions that mimic artifacts in blood smear microscopy and  acute myeloid leukemia, including noise and blurring effects.}
    \label{fig:cell_microscopy}
\end{figure*}

\begin{figure*}[t]
    \centering
    \includegraphics[width=1.0\textwidth]{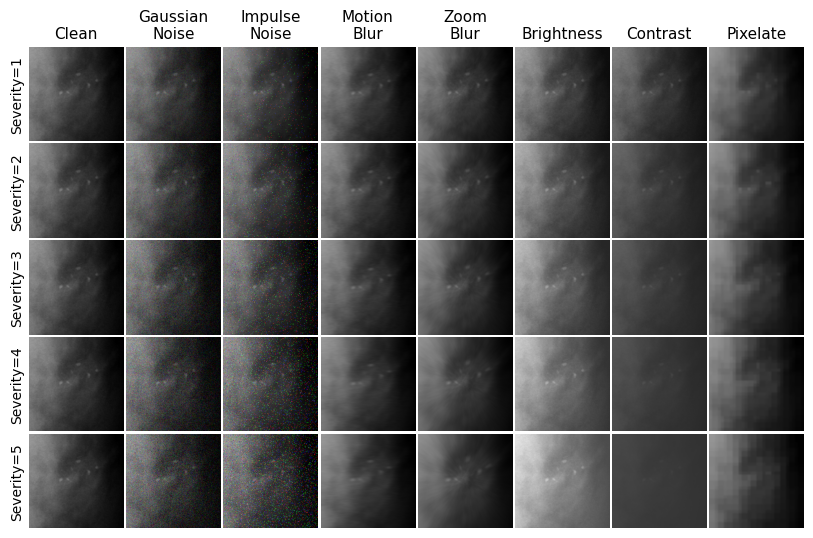}
    \vspace{-0.8cm}
    \caption{\small Example images of Breast Imaging Scans including \textbf{MAMMO-C} from MediMeta-C, showcasing different corruption types. These corruptions simulate real-world degradation in mammography calcification scans.}
    \label{fig:mammo}
    \vspace{-0.5cm}
\end{figure*}

\begin{figure*}[t]
    \centering
    \includegraphics[width=1.0\textwidth]{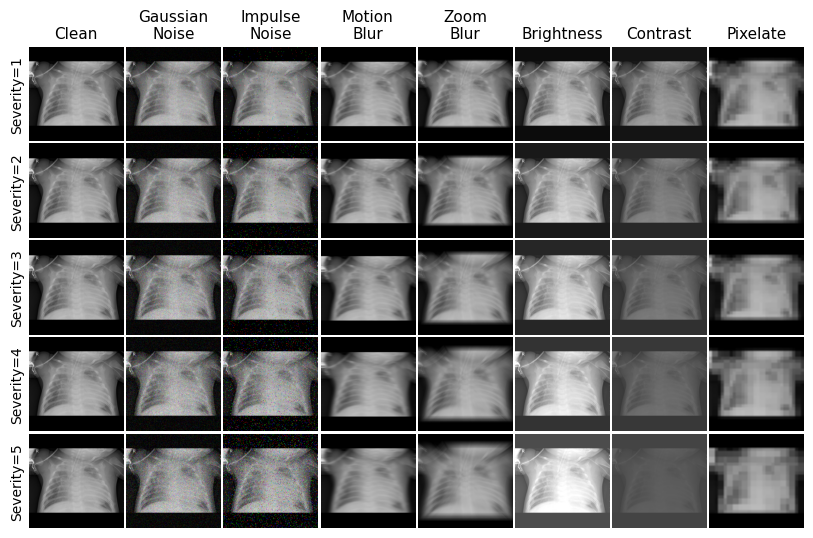}
    \vspace{-0.8cm}
    \caption{\small Example images from \textbf{PNEUMONIA-C} in MediMeta-C, demonstrating corruption types commonly encountered in chest X-ray scans, such as motion blur and pixelation.}
    \label{fig:pneumonia}
\end{figure*}

\begin{figure*}[t]
    \centering
    \includegraphics[width=1.0\textwidth]{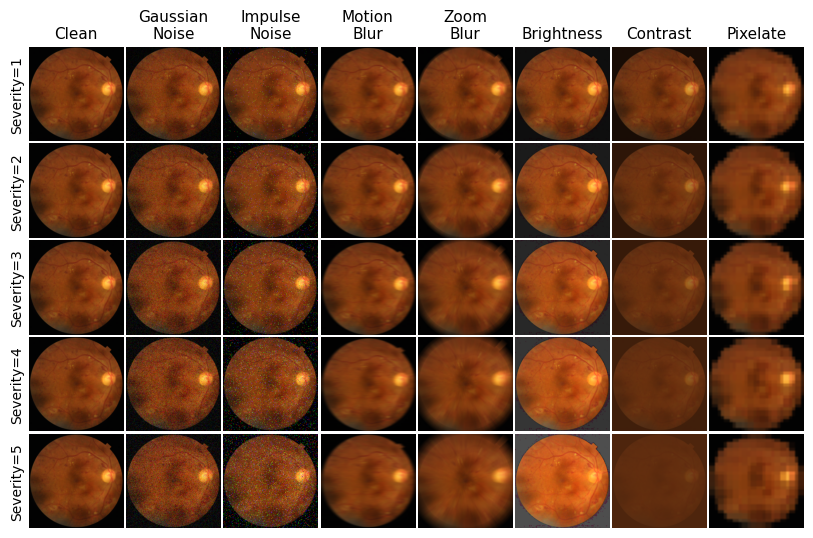}
    \vspace{-0.8cm}
    \caption{\small Example images from \textbf{FUNDUS-C} in MediMeta-C, displaying distortions of Retinal Fundus scans that replicate issues in Fundoscopic examination, such as sensor noise and defocus blur.}
    \label{fig:fundus}
    \vspace{-0.5cm}
\end{figure*}

\begin{figure*}[t]
    \centering
    \includegraphics[width=1.0\textwidth]{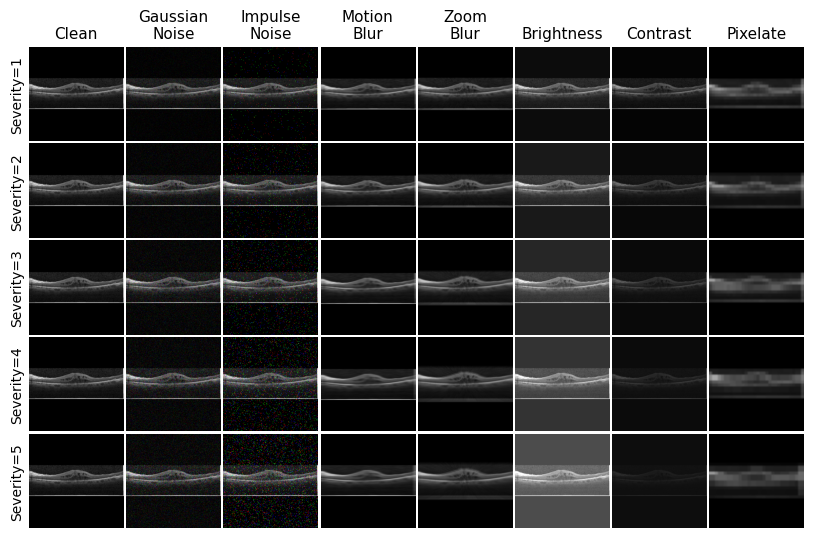}
    \vspace{-0.8cm}
    \caption{\small Example images from \textbf{OCT-C} in MediMeta-C, displaying distortions of Retinal OCT scans that replicate issues in Optical Coherence Tomography (OCT) imaging, such as sensor noise and defocus blur.}
    \label{fig:oct}
\end{figure*}

\end{document}